\documentclass[journal]{IEEEtran}
\IEEEoverridecommandlockouts
\usepackage{cite}
\usepackage{amsmath,amssymb,amsfonts}
\usepackage{subfig}
\usepackage[normalem]{ulem}
\usepackage{algorithmic}
\usepackage{graphicx}
\usepackage{textcomp}
\usepackage{amsbsy}
\usepackage{xcolor}

\usepackage{xspace}

\DeclareMathOperator{\argmax}{argmax}

\newcommand{\bx}{\mathbf{x}}
\newcommand{\by}{\mathbf{y}}
\newcommand{\bz}{\mathbf{z}}
\newcommand{\bw}{\mathbf{w}}
\newcommand{\br}{\mathbf{r}}
\newcommand{\btheta}{{\boldsymbol\theta}}
\newcommand{\bdelta}{{\boldsymbol\delta}}
\newcommand{\R}{\mathbb{R}}
\newcommand{\C}{\mathbb{C}}
\renewcommand{\S}{\mathcal{S}}
\newcommand{\I}[1]{\mathbb{I}\{#1\}}
\newcommand{\DMS}{\emph{DMS}}
\newcommand{\BDMS}{\emph{BDMS}\xspace}
\newcommand{\BODMS}{\emph{BODMS}\xspace}
\newcommand{\Oracle}{\emph{Oracle}\xspace}
\newcommand{\PDMS}{\emph{PDMS}\xspace}
\newcommand{\RNI}{\emph{RNI}\xspace}
\newcommand{\BBDMS}{\emph{BB-DMS}\xspace}
\newcommand{\GFSK}{\textsc{GFSK}\xspace}
\newcommand{\CPFSK}{\textsc{CPFSK}\xspace}
\newcommand{\PSK}[1]{\textsc{PSK#1}\xspace}
\newcommand{\BPSK} {\textsc{BPSK}\xspace}
\newcommand{\QPSK} {\textsc{QPSK}\xspace}
\newcommand{\PAM}[1]{\textsc{PAM#1}\xspace}
\newcommand{\QAM}[1]{\textsc{QAM#1}\xspace}
\newcommand{\NoPerturb}{\emph{NoPerturb}\xspace}
\newcommand{\Baseline}{\emph{Baseline}\xspace}

\newcommand{\graphwidth}{0.34\textwidth}

\begin{document}

\title{The Best Defense Is a Good Offense: Adversarial Attacks to Avoid Modulation Detection
\thanks{This work was presented in part at the 7th IEEE Global Conference on Signal and Information Processing (GlobalSIP 2019) \cite{hameed2019communicationGS}.}
\thanks{This work was supported by an Imperial College London President's PhD Scholarship, and by the European Research Council (ERC) through the Starting Grant BEACON (No. 677854).}
}

\author{\IEEEauthorblockN{Muhammad Zaid Hameed$^1$,  Andr\'as Gy\"orgy$^{2}$, and Deniz G\"und\"uz$^1$\\}
\smallskip
\IEEEauthorblockA{
$^1$Department of Electrical and Electronic Engineering,
Imperial College London, UK \\
$^2$DeepMind, London, UK \\
Email: muhammad.hameed13@imperial.ac.uk, agyorgy@google.com, d.gunduz@imperial.ac.uk
}}

\maketitle

\begin{abstract}
We consider a communication scenario, in which an intruder
tries to determine the modulation scheme of the intercepted signal. Our aim is to minimize the accuracy of the intruder, while guaranteeing that the intended receiver can still recover the underlying message with the highest reliability. This is achieved by perturbing channel input symbols at the encoder, similarly to adversarial attacks against
classifiers in machine learning. In
image classification, the perturbation is limited to be imperceptible to a human observer, while in our case the perturbation is constrained so that the message can still be reliably decoded by the legitimate receiver, which is oblivious to the perturbation. Simulation results demonstrate the viability of our approach to make wireless communication secure against
state-of-the-art intruders (using deep learning or decision trees) with minimal sacrifice in the communication performance. On the other hand, we also demonstrate that using diverse training data and curriculum learning can significantly boost the accuracy of the intruder.
\end{abstract}

\begin{IEEEkeywords}
secure communication, deep learning, adversarial attacks, modulation classification
\end{IEEEkeywords}

\section {Introduction}

Securing wireless communication links is as essential as increasing their efficiency and reliability, for military, commercial, as well as consumer communication systems. The standard approach to securing communications is to encrypt the transmitted data. However, encryption may not always provide full security (e.g., in case of side-channel attacks), or strong encryption may not be available due to complexity limitations (e.g., for IoT devices). To further improve security, encryption can be complemented with other techniques, preventing the adversary from even recovering the encrypted bits.

As outlined in \cite{Prescott1993PerformanceMF}, an adversary implements its attacks on a wireless communication link in four steps: 1) tunes into the frequency of the transmitted signal; 2) detects whether there is signal or not; 3) intercepts the signal by extracting its features; and 4) demodulates the signal by exploiting the extracted features, and obtains a binary stream of data. Preventing any of these steps can strengthen the security of the communication link. While encryption focuses on protecting the demodulated bit stream, physical layer security \cite{Wyner:BSTJ:75, Gunduz:security_feedback} targets the fourth step by minimizing the mutual information available to the intruder. Recently, there has also been significant interest in preventing the second step through covert communications \cite{Bash:arxiv:12}. In this work, we instead focus on the third step, and aim at preventing the adversary from detecting the modulation scheme used for communications.

Modulation detection is the step between signal detection and demodulation in communication systems, and thus plays an important role in data transmission, as well as in detection and jamming of unwanted signals in military communications and other sensitive applications  \cite{dobre2007survey}.
Recently, deep learning techniques have led to significant progress in modulation-detection accuracy:  methods based on convolutional and other deep neural networks can detect the modulation scheme directly from raw time-domain samples \cite{mendis2016deep,o2017introduction, west2017deep, kim2016deep, liu2017deep}, surpassing the accuracy of conventional modulation detectors based on likelihood function or feature-based representations (see \cite{dobre2007survey} for a survey of these approaches).

Our aim in this paper is to prevent an intruder that employs a state-of-the-art modulation detector from successfully identifying the modulation scheme being used. The rationale behind this is that if the intruder is unable to identify the modulation scheme, it is unlikely to be able to decode the underlying information or employ modulation-dependent jamming techniques to prevent communication. To achieve this goal, we introduce modifications to the transmitted signal. The main challenge here is to guarantee that the intended receiver of the (modified) transmitted signal can continue to receive the underlying message reliably, while preventing the intruder from detecting the modulation scheme being used. Otherwise, reducing the accuracy of the modulation-detecting intruder would be trivial by sacrificing the performance of the intended receiver. We assume that the intended receiver is oblivious to the modifications employed by the transmitter to confuse the intruder, and, therefore, the goal of the transmitter is to introduce as small modifications to the transmitted signal as possible that are sufficient to fool the intruder but not larger than the error correction capabilities of the intended receiver.

Introducing small variations into the modulation scheme that can fool 
an intruder is similar to adversarial attacks on classifiers, in particular, deep neural networks (DNNs) \cite{szegedy2013intriguing, goodfellow2014explaining}. In the literature, adversarial attacks are mostly considered in the area of image classification, where they pose security risks  by exposing the vulnerabilities of classifiers against very small changes in the input that are imperceptible to humans but lead to incorrect decisions. In contrast, we exploit the same approach here to defend a communication link against an intruder that employs DNNs or other standard classification methods for interception.

In \cite{sadeghi2018adversarial}, an adversarial attack for a deep-learning-based modulation classifier has been proposed where the adversary assumes the availability of noisy symbols received  at the modulation classifier for generating the adversarial attack, which makes it impractical and limited in scope. A similar method has been proposed  recently in \cite{kokalj2019mitigation}, where modifications are employed by the transmitter to evade a DNN-based jammer, and the receiver uses another DNN (an autoencoder) to preprocess the received signal and filter out the modifications. However,
no analysis has been provided on the impact of this method
on the bit error rate (BER) of the received signal. In contrast to \cite{kokalj2019mitigation}, we do not limit our approach to a DNN-based jammer and consider a
receiver that is completely oblivious to the modifications in the transmitter.

We also consider the impact of the
defensive perturbations on the BER at the legitimate receiver.
The results of \cite{kokalj2019mitigation} has been extended in \cite{kokalj2019targeted} to the detection of wireless communication protocols, and targeted adversarial attacks are also considered to generate the perturbations.\footnote{Targeted adversarial attacks \cite{carlini2017towards} aim to modify the data so that the attacked classifier predicts an incorrect class selected by the attacker.}

A number of concurrent works have also appeared in the literature (parallel or following the original publication of our preprint at arXiv \cite{hameed2019communication} and the conference version of our paper \cite{hameed2019communicationGS}). Most similar to ours is \cite{flowers2019communications}, which proposes modifications in the transmitted signal using an adversarial residual network at the transmitter to evade the modulation detector at an intruder while the legitimate receiver is able to decode the signal with small bit error rate. Compared to this paper, we use different adversarial attack techniques, propose different ways of improving the modulation-detection accuracy of the intruder, and analyze the trade-off between the code rate and the BER for defensive perturbations and an improved intruder.

Adversarial perturbations have also been applied to attack a legitimate receiver in \cite{flowers2019evaluating,kim2020overtheair}. In these works the input signal of the receiver is perturbed by an over-the-air attack to make the modulation classifier of the legitimate receiver fail (cf. in our case the transmitter changes the signal to fool the intruder). In \cite{flowers2019evaluating}, the performance of such attacks is evaluated in terms of the BER at the receiver under the impractical assumption that the attacker has full knowledge of the signal at the receiver. The same scenario has been considered in \cite{kim2020overtheair}, and attack methods of various strength have been devised under more realistic assumptions about the capabilities of the attacker, in particular about its information on the signal received by the modulation classifier (fully known vs. its distribution being estimated based on samples available to the attacker) and on the channel noise from the attacker to the receiver (knowing the exact realization or just the noise distribution). While these attack methods share the underlying idea with our defensive perturbations, they face a much easier problem, as the attacks are not constrained by ensuring a low BER at a distinct receiver.

While we consider adversarial attack methods that affect the behavior of trained classifiers (i.e., the modulation classifier of the intruder in our case) by perturbing their input data (these attacks are known as test-time or evasion attacks), another class of adversarial machine learning algorithms, called poisoning attacks, aim to compromise the training procedure of classifiers and other machine learning models by modifying their training data \cite{barreno2010taxonomy}. Poisoning attacks have been used in launching and avoiding jamming attacks in wireless communication \cite{sagduyu2019iot, shi2018spectrum, erpek2018deep}; however, since these methods address the training of the machine learning models employed by the jammer and the transmitter, they are orthogonal to our developments.

In summary, our main contributions are as follows:
\begin{itemize}
    \item We propose a novel defense mechanism that modifies the channel input symbols at the transmitter in order to reduce the modulation-classification accuracy at the intruder while maintaining a low BER at the legitimate receiver.
    \item We provide a thorough experimental evaluation of the effect of these modifications on the BER of different modulation schemes.
    \item We demonstrate that by using training data obtained from different SNR values and employing curriculum learning, an intruder can learn a classifier that is much more robust against both the channel noise and the defensive perturbations, improving upon the state of the art in our experiments when no defense mechanism is applied.
    \item We show that by reducing the communication rate, the BER at the legitimate receiver can be reduced  while the intruder is limited to achieve the same or worse modulation-classification accuracy.
\end{itemize}

The rest of the paper is organized as follows: The system model is described in Section~\ref{sec:sys_model}, followed by the description of our novel modulation perturbation methods in Section~\ref{sec:perturbation}. Experimental results are presented in Section~\ref{sec:exp_eval}, while conclusions are drawn and future work is discussed in Section~\ref{sec:conclusions}.

\section{System Model}\label{sec:sys_model}

Consider a transmitter that maps a binary input sequence $\bw \in \{0, 1\}^m$ into a sequence of $n$ complex channel input symbols, $\bx \in \C^n$, employing forward error correction coding. The input data is first encoded by the channel encoder, and then modulated for transmission.
Formally, the modulated signal $\bx$ is obtained as $\bx=M_s(\bw)$, where $s \in \S$ is the employed modulation scheme with $\S$ denoting the finite set of available modulation schemes, and for any $s$, $M_s:\{0,1\}^m \to \C^n$ denotes the whole encoder function with modulation $s$. We assume that $M_s$ satisfies the power constraint $(1/n) \| \bx \|_2^2 \le 1$ for any input sequence $\bw$. After encoding, signal $\bx$ is sent over a noisy channel, assumed to be an additive white Gaussian noise (AWGN) channel for simplicity: baseband signals $\by_1$ and $\by_2$, received by the intended receiver and the intruder, respectively, are given by
\begin{align}
\by_i = M_s(\bw) + \bz_i = \bx + \bz_i,\; i=1,2,
\end{align}
where $\bz_1,\bz_2 \in \C^n$ are independent channel noise (also independent of $\bx$) with independent zero-mean complex Gaussian components with variance $\sigma_1^2$ and $\sigma_2^2$, respectively.

The intended receiver, upon receiving the sequence of noisy channel symbols ${\mathbf{y_{1}}}$, demodulates the received signal, and decodes the underlying message bits with the goal of minimizing the (expected) BER $\mathbb{E[e(\mathbf{w}, \mathbf{y_{1}})]}$, where
\begin{equation}
\label{eq:BER}
\textstyle
e(\mathbf{w},\by_1) \triangleq \frac{1}{m}\sum_{i=1}^m \I{w_i\neq \hat{w}_i},
\end{equation}
$\mathbf{\hat{w}}$ is the decoded bit sequence from $\by_1$, and the expectation is over the uniformly random input bit sequence $\mathbf{w}$ and the noise sequence $\mathbf{z_{1}}$.\footnote{For any event $E$, $\I{E}=1$ if $E$ holds, and $0$ otherwise. Furthermore, for any real or complex vector $\mathbf{v}$, $v_i$ denotes its $i$th coordinate.}

The intruder aims to determine the modulation scheme employed by the transmitter based on its received noisy channel output $\by_2$. The transmitter, on the other hand, wants to communicate without its modulation scheme being correctly detected by the intruder, while keeping the BER at an acceptable level.

Formally, the aim of the intruder is to determine, for any sequence of channel output symbols $\by_2 \in \C^n$, the modulation method used by the transmitter. This leads to a \emph{classification} problem where the label $s \in \S$ is the employed modulation scheme, and the input to the classifier is the received channel sequence $\by_2 \in \C^n$. We consider the case in which the intruder implements a score-based classifier, and assigns to $\by_2$ the label $\hat{s}=\argmax_{s' \in \S} f_\btheta(\by_2,s')$, where $f_\btheta: \C^n \times \S \to \R$ is a score function  parametrized by $\btheta \in \R^d$, which assigns a score (pseudo-likelihood) to each possible class $s' \in \S$ for every $\by_2$, and finally selects the class with the largest score. With a slight abuse of notation, we denote the resulting class label by $\hat{s} = f_\btheta(\by_2)$.
The goal of the intruder is to maximize the probability $\Pr(s = \hat{s})$ of correctly detecting the modulation scheme, which we will also refer to as the success probability of the intruder.\footnote{Here we assume an underlying probabilistic model about how the the bit sequence \textbf{w} and modulation scheme is selected.}
For state-of-the-art modulation detection schemes \cite{mendis2016deep,o2017introduction, west2017deep, kim2016deep, liu2017deep}, $f_{\btheta}$ is a convolutional neural network classifier, $\btheta$ is the vector of the weights of the neural network, while the $f_\btheta(\by_2,s')$ are the so-called logit values for the class labels $s' \in \S$.

The performance of both the intended receiver, measured by the BER, and the intruder, measured by the detection accuracy, depend on the signal-to-noise ratio (SNR) of the corresponding channels, $\frac{1}{\sigma_{1}^2}$ and $\frac{1}{\sigma_{2}^2}$, respectively. We assume that these SNR values are known by the legitimate receiver and the intruder, which can employ a specific $f_\btheta$ for each SNR value. We will also assume that the intruder has access to training data at the SNR value $\frac{1}{\sigma_{2}^2}$ to train $f_{\btheta}$. This can be done offline as the intruder can generate as much training data as required at a specific SNR value.

\section{Modulation Perturbation to Avoid Detection}\label{sec:perturbation}

In this paper we intend to modify the encoding processes $M_s$ such that, given a modulation scheme $s \in \S$, the new encoding method $M'_s$ ensures that the intruder's success probability gets smaller, while the BER of the receiver (using the same decoding procedure for $M_s$) does not increase substantially. Our solution is motivated by adversarial attacks for image classification, where it is possible to modify images such that the modification is imperceptible to a human observer, but it makes state-of-the-art image classifiers to err \cite{szegedy2013intriguing, goodfellow2014explaining}. Adversarial examples are particularly successful in fooling high-dimensional DNN classifiers. Applying the same idea to our problem, we aim to find defensive modulation schemes $M'_s$ such that $M'_s(\bw) \approx M_s(\bw)$, but the intruder misclassifies the new received signal $\by'_2=M'_s(\bw)+\bz_2$ with higher probability.

\subsection{Adversarial attack in an idealized scenario}\label{sec:Ideal}

Following directly the idea of adversarial attacks on image classifiers \cite{goodfellow2014explaining}, an idealized yet impractical adversarial attack mechanism is proposed in \cite{sadeghi2018adversarial} which modifies a correctly classified channel output sequence $\by_2$ (i.e., for which $s=f_\btheta(\by_2)$) with a perturbation $\bdelta \in \C^n$ such that $f_\btheta(\by_2+\bdelta) \neq f_\btheta(\by_2)$, the true label, while imposing the restriction $\| \bdelta \|_2 \le \epsilon$ for some small positive constant $\epsilon$.
Thus, to mask the modulation scheme, the goal is to find, for each correctly classified $\by_2$ separately, a perturbation $\bdelta$ that maximizes the zero-one loss:
\begin{equation}
\label{eq:adversarial}
\begin{split}
    & \text{maximize }\; \I{f_\btheta(\by_2+\bdelta) \neq s}  \text{ such that }\; \|\bdelta\|_2 \le \epsilon~,
    \end{split}
\end{equation}
where $s=f_\btheta(\by_2)$ is the true modulation label.

If the maximum is $1$, such a $\bdelta$ results in a successful adversarial perturbation and a successful adversarial example $\by_2+\bdelta$ (i.e., one for which the intruder makes a mistake).
This approach, however, has two limitations. First of all, as opposed to image classifiers, we are not concerned with the visual similarity of the perturbed signal $\by_{2} + \bdelta$ to the original one, $\by_{2}$. The reason for bounding the perturbation $\bdelta$ is instead to guarantee that the BER at the intended receiver is still limited. Moreover, in practice we do not have access to $\by_2$, as it does not only depend on $\bx$, but also on the channel noise $\bz_2$, which is not available at the transmitter. Therefore, the above mechanism, analyzed in \cite{sadeghi2018adversarial}, is an \emph{oracle} scheme working under some idealized assumptions, and we use it only as a baseline.

It remains to give an algorithm that finds an adversarial perturbation $\bdelta$ solving problem \eqref{eq:adversarial}. However, we note that the target function $\I{f_\btheta(\by_2+\bdelta) \neq f_\btheta(\by_2)}$ is binary, and so no gradient-based search is directly possible. To alleviate this, usually a surrogate loss function $L(\btheta,\by_2,s)$ to the zero-one loss is used (which is often also used in training the classifier $f_\btheta$), which is amenable to gradient-based (first-order) optimization. For classification problems, a standard choice is the cross-entropy loss defined as $L(\btheta,\by_2,s) = -\log(1+e^{-f_\btheta(\by_2,s)})$, and one can search for adversarial perturbations by solving
\begin{align}
        \label{eq:advL}
        \text{maximize }\; L(\btheta,\by_2+\bdelta,s)
        \text{ such that }\; \|\bdelta\|_2 \le \epsilon. 
\end{align}
\vspace{-0.5cm}

Different methods are used in the literature to solve \eqref{eq:advL} approximately \cite{goodfellow2014explaining,kurakin2016adversarial, chen2017ead, carlini2017towards}.
In this paper we use the state-of-the-art projected (normalized) gradient descent (PGD) attack \cite{madry2017towards} to generate adversarial examples, which is an iterative method: starting from $
\by^0=\by_2$, at each iteration $t$ it calculates
\begin{equation}
\label{eq:attack_iter}
\by^{t} =  \Pi_{\mathcal{B}_{\epsilon}(\by_2)}\big( \by^{t-1} + \beta \ \mathrm{sign}(\nabla_{\!\by} L(\btheta, \by^{t-1},s)) \big),
\end{equation}
where $\beta>0$ denotes the step size, $\mathrm{sign}$ denotes the sign operation, and $\Pi_{\mathcal{B}_{\epsilon}(\by_2)}$ denotes the Euclidean projection operator to the $L_2$-ball $\mathcal{B}_{\epsilon}(\by_2)$ of radius $\epsilon$ centered at $\by_2$, while $\nabla$ denotes the gradient. The attack is typically run for a specified number of steps, which depends on the computational resources; in practice $\by^t$ is more likely to be a successful adversarial example for larger values of $t$. We will refer to this \emph{idealized} modulation scheme as the \emph{Oracle Scheme} (\Oracle).

Note that this formulation assumes that we have access to the logit function $f_\btheta$ of the intruder; these methods are called \emph{white-box} attacks. If $f_\btheta$ is not known, one can create adversarial examples against another classifier $f_{\btheta'}$, and hope that it will also work against the targeted model $f_\btheta$. Such methods are called \emph{black-box} attacks, and are surprisingly successful against image classifiers \cite{papernot2017practical}. We will also consider black-box attacks against intruders in our experimental evaluations.

\subsection{Adversarial attack through channel input modification}\label{sec: practical_methods}

As mentioned before, the \Oracle scheme is infeasible in practice as the transmitter can only modify the channel input $\bx=M_s(\bw)$ but not $\by_2$ directly. Thus, the new modulation scheme is defined as

\begin{equation}
M'_s(\bw)=\alpha (M_s(\bw)+\bdelta),
\end{equation}

\noindent where we will consider different choices for $\bdelta \in \C^n$, and the multiplier $\alpha = \sqrt{n}/\|M_s(\bw)+\bdelta\|_2$ is used to ensure that the new channel input $\bar{\bx} = M'_s(\bw)$ satisfies the average power constraint $(1/n) \| \bar{\bx}\|_2^2 \le 1$. The signals received at the receiver and at the intruder are $\bar{\by}_1 = \bar{\bx} + \bz_1$ and $\bar{\by}_2 = \bar{\bx} + \bz_2$, respectively.
The difficulty in this scenario is that the effect of any carefully designed perturbation $\bdelta$ may be (and, in fact, is in practice) at least partially masked by the channel noise. Furthermore, since now the perturbed signal is transmitted at the actual SNR of the channel, the effective SNR of the system is decreased, as the transmitted signal already includes the perturbation $\bdelta$, which can be treated as noise from the intended receiver's point of view.

Our first and simplest method to find a perturbation $\bdelta$  disregards the effects of the channel noise and the resulting BER at the receiver.

\subsubsection{Perturbation-based Defensive Modulation Scheme (PDMS)} In this method, called the \PDMS, we aim to solve the optimization problem \eqref{eq:advL} with $\bx$ in place of $\by_2$, via \eqref{eq:attack_iter} initialized at $\by^0=\bx$ and with projection to $\mathcal{B}_\epsilon(\bx)$ (for a specified number of iterations $t$ and perturbation size $\epsilon$).

\subsection{BER-aware adversarial attack}
Next, we consider methods that also take into account the BER, $e(\bar{\by}_1,\bw)$ at the receiver (see Eq.~\ref{eq:BER}): that is, instead of enforcing the perturbation $\bdelta$ to be small and hoping for only a slight increase in the BER, we optimize also for the latter. There is an inherent trade-off between these two targets: a larger $\bdelta$ results in a bigger reduction in the detection accuracy of the intruder, but will also increase the BER at the receiver. We consider two methods to handle this trade-off:

In the first one, called \emph{BER-Aware Defensive Modulation Scheme (BDMS)};
we consider a (signed) linear combination of our two target functions in order to balance the above two effects,
\begin{align*}
L_\lambda(\btheta,\bar{\bx},s,\bz_1,\bz_2)= L(\btheta,\bar{\bx}+\bz_2,\bdelta) - \lambda e(\bar{\bx}+\bz_1,\bw)
\end{align*}
for some $\lambda>0$, where $\bar{\by}_i=\bar{\bx}+\bz_i, i=1,2$,
and aim to find a perturbation $\bdelta$ or, equivalently, a modulated signal $\bar{\bx}=\bx+\bdelta$ that maximizes the expectation
\begin{equation}
\label{eq:combined}
\mathbb{E}_{\bz_1,\bz_2} [L_\lambda(\btheta,\bx,s,\bz_1,\bz_2)]
\end{equation}
with respect to the channel noise $\bz_1,\bz_2$.
Here we can use stochastic gradient ascent\footnote{However, similarly to the literature on adversarial attack methods, we often call these methods gradient \emph{descent} instead of ascent.} to compute an approximate local optimum, but in practice we find that enforcing $\bdelta$ to be small during iterations improves the performance; hence, we use a stochastic version of PGD optimization \eqref{eq:attack_iter}: starting at $\bx^0=\bx$, our candidate for $\bar{\bx}$ is iteratively updated as
\[
\bx^t=\Pi_{\mathcal{B}_{\epsilon}(\bx)}\big(\bx^{t-1} + \beta \cdot \mathrm{sign} ( \nabla_{\!\bx}L(\btheta,\bx^{t-1},s,\bz^t_1,\bz^t_2)) \big),
\]
where $\bz^t_i$ are independent copies of $\bz_i$, respectively, for $i=1,2,$ and $t=1,2,\ldots$. Although $\mathbb{E}_{\bz_1} [e(\bar{\bx}+\bz_1,\bw)]$ is differentiable, $e(\by,\bw)$ for a given fixed value of $\by$ is not (since it takes values from the finite set $\{0,1/n,\ldots,1\}$). Similarly to \cite{uesato2018adversarial}, we approximate the gradient of the expected error using simultaneous perturbation stochastic approximation (SPSA) \cite{spall1992multivariate} as
\begin{align}
    \label{eq:SPSA}
{\nabla}_{\!\by}\, e(\by,\!\bw)
\!=\! \frac{1}{K}\sum_{k=1}^K \frac{e(\by\!+\!\eta \br_k,\bw)-e(\by\!-\!\eta \br_k,\bw)}{2\eta} \br_k^\top,
\end{align}
where $\br_1,\ldots,\br_K$ are random vectors selected independently and uniformly from $\{-1,1\}^n$.

In the alternative \emph{BER-Aware Orthogonal Defensive Modulation Scheme (BODMS)},
instead of maximizing the combined target \eqref{eq:combined}, we try to maximize the cross-entropy loss $L(\btheta,\bar{\by}_2,s)$ while not increasing (substantially) the BER $e(\bar{\by}_1,\bw)$. In order to do so, we maximize $L(\btheta,\bar{\by}_2,s)$ using stochastic PGD (again, in every step we choose independent noise realizations), but we restrict the steps in the directions where the BER does not change. Thus, in every step we update $\bx^{t-1}$ in a direction \emph{orthogonal} to the gradient of the BER defined as
\begin{align*}
\lefteqn{\nabla_{\!o} L(\btheta,\bx^{t-1}+\bz^t_2,s) } \\
& \triangleq  \nabla_{\!\bx} L(\btheta,\bx^{t-1}+\bz^t_2,s) -
\bigl\langle \nabla_{\!\bx} L(\btheta,\bx^{t-1}+\bz^t_2,s), d_{e}
\bigr\rangle d_{e}
\end{align*}
where
$d_{e}= {\nabla}_{\!\bx} e(\bx^{t-1}+\bz^t_1,\bw)/\|{\nabla}_{\!\bx} e(\bx^{t-1}+\bz^t_1,\bw)\|_2$ is the (approximate) gradient direction of the BER (computed, e.g., using SPSA as in Eq.~\ref{eq:SPSA}).

\section{Experimental Evaluation}\label{sec:exp_eval}

In this section, we test and compare the performance of the proposed methods through numerical simulations. We assume that the binary source data is generated independently and uniformly at random, and is encoded using a rate 2/3 convolutional code before modulation. Eight standard baseband modulation schemes are considered: \GFSK, \CPFSK, \PSK8, \BPSK, \QPSK, \PAM4, \QAM{16}, \QAM{64}. A square-root raised cosine filter is used for pulse shaping of the modulated data with a filter span of 10,  roll-off factor of 0.25 and upsampling factor of 8 samples per symbol and the modulated data is sent over an AWGN channel with SNR varying between -20 dB and 20 dB. We consider identical SNRs during both the training of the intruder and at test time. After hard decision demodulation, the receiver uses Viterbi decoding to estimate the original source data.

We follow the setup of \cite{o2017introduction} for modulation detection: The intruder has to estimate the modulation scheme after receiving 128 complex I/Q (in-phase /quadrature) channel symbols;  this is because we assume that the modulation detection is only the first step for the intruder, which then uses this information for either trying to decode the message or to interfere with its transmission. Therefore, the modulation detection should be completed based on a short sequence of channel symbols. As the classifier, we first consider the deep convolutional neural network architecture of \cite{o2017introduction} for the intruder
which operates on the aforementioned 256-dimensional data.

\begin{figure}[t]
  \centering
  \includegraphics[width=\graphwidth]{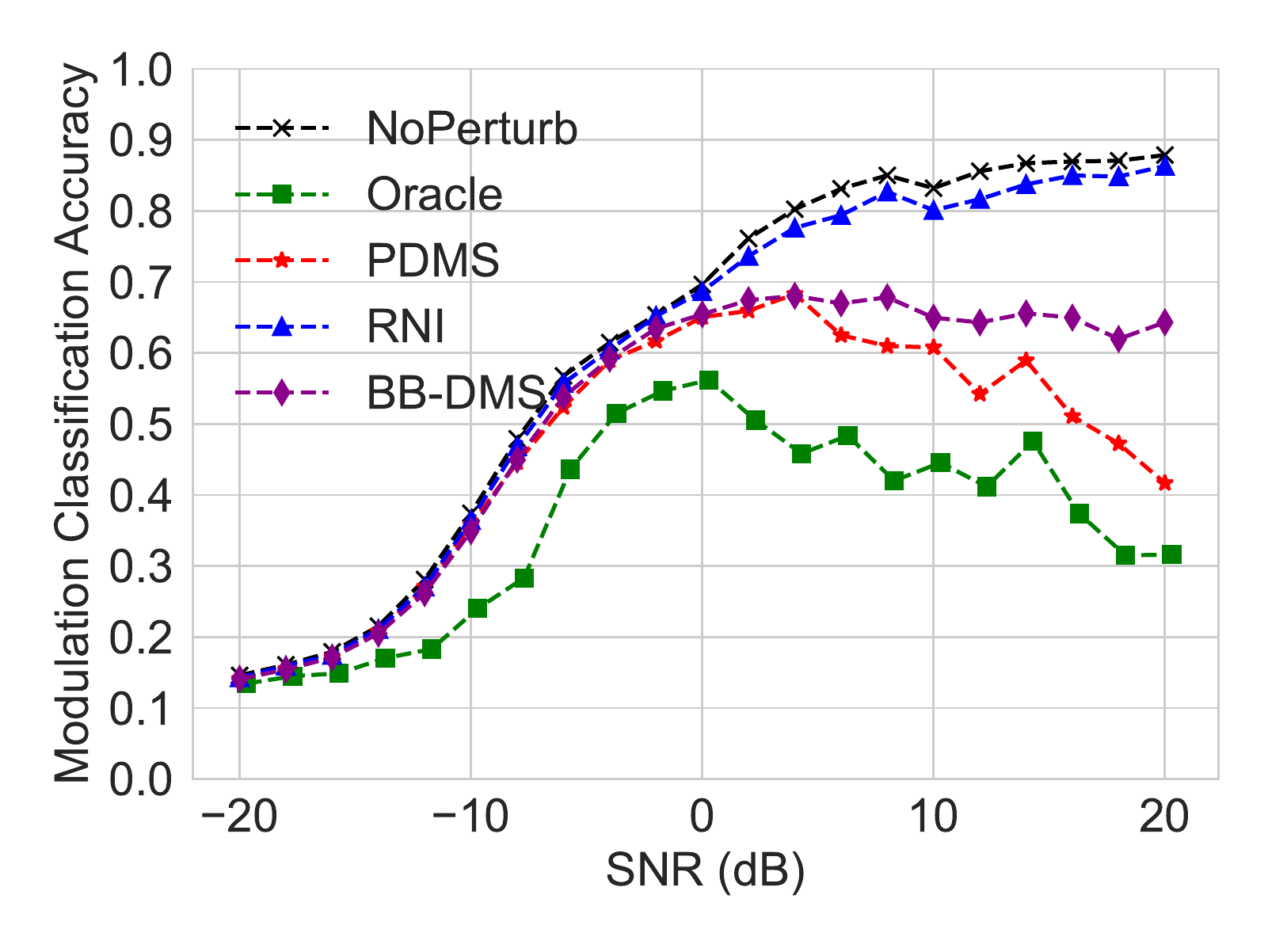}
  \caption{Modulation-classification accuracy of the intruder as a function of SNR for different defensive modulation schemes.}
  \label{fig: accu_cr_2_by_3}
   \subfloat[\PSK8  \label{subfig-1:PSK8_2_by_3}]{%
       \includegraphics[width=\graphwidth]{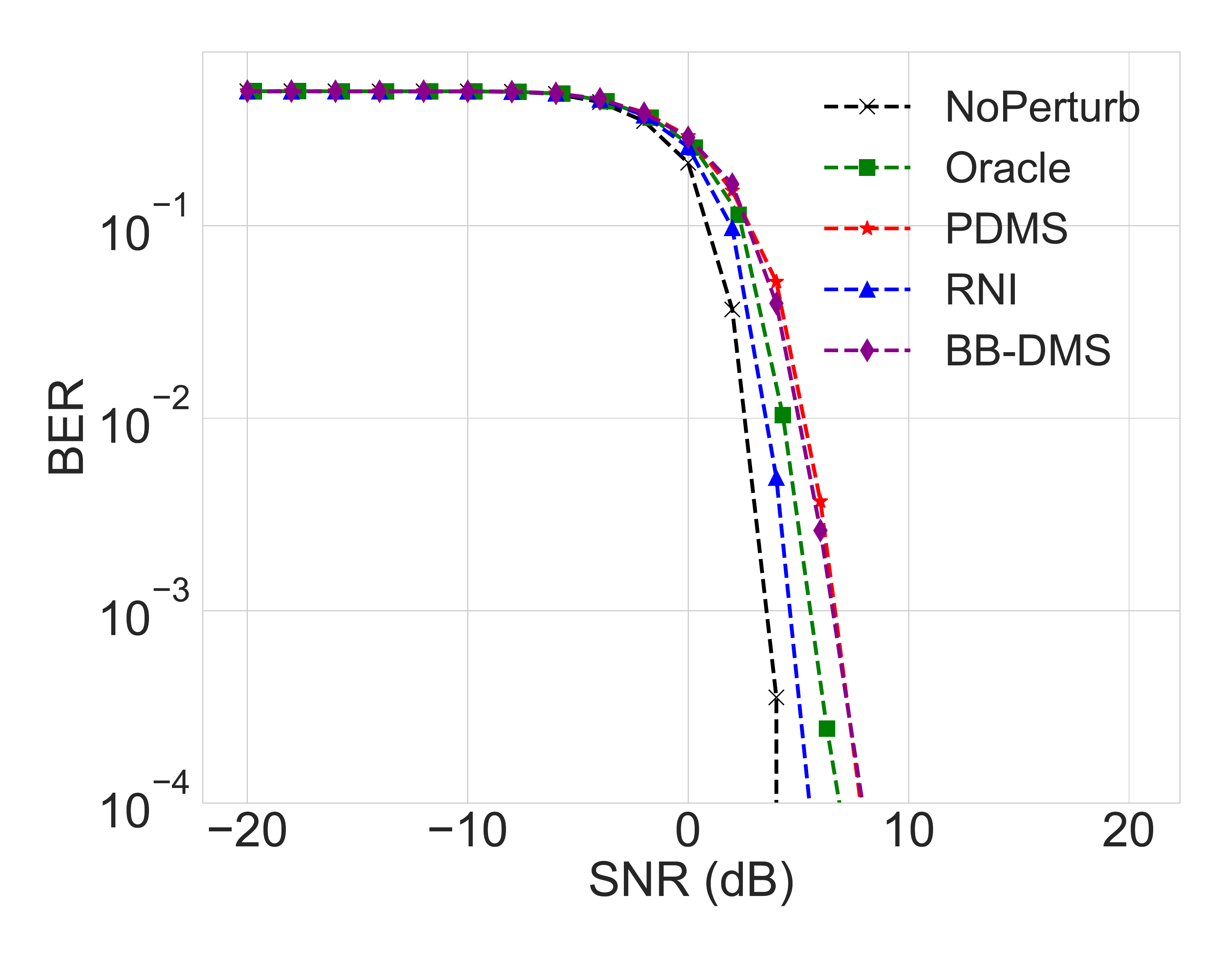}
     }
     \hfill
     \subfloat[\QAM{64} \label{subfig-2:qam64_2_by_3}]{%
       \includegraphics[width=\graphwidth]{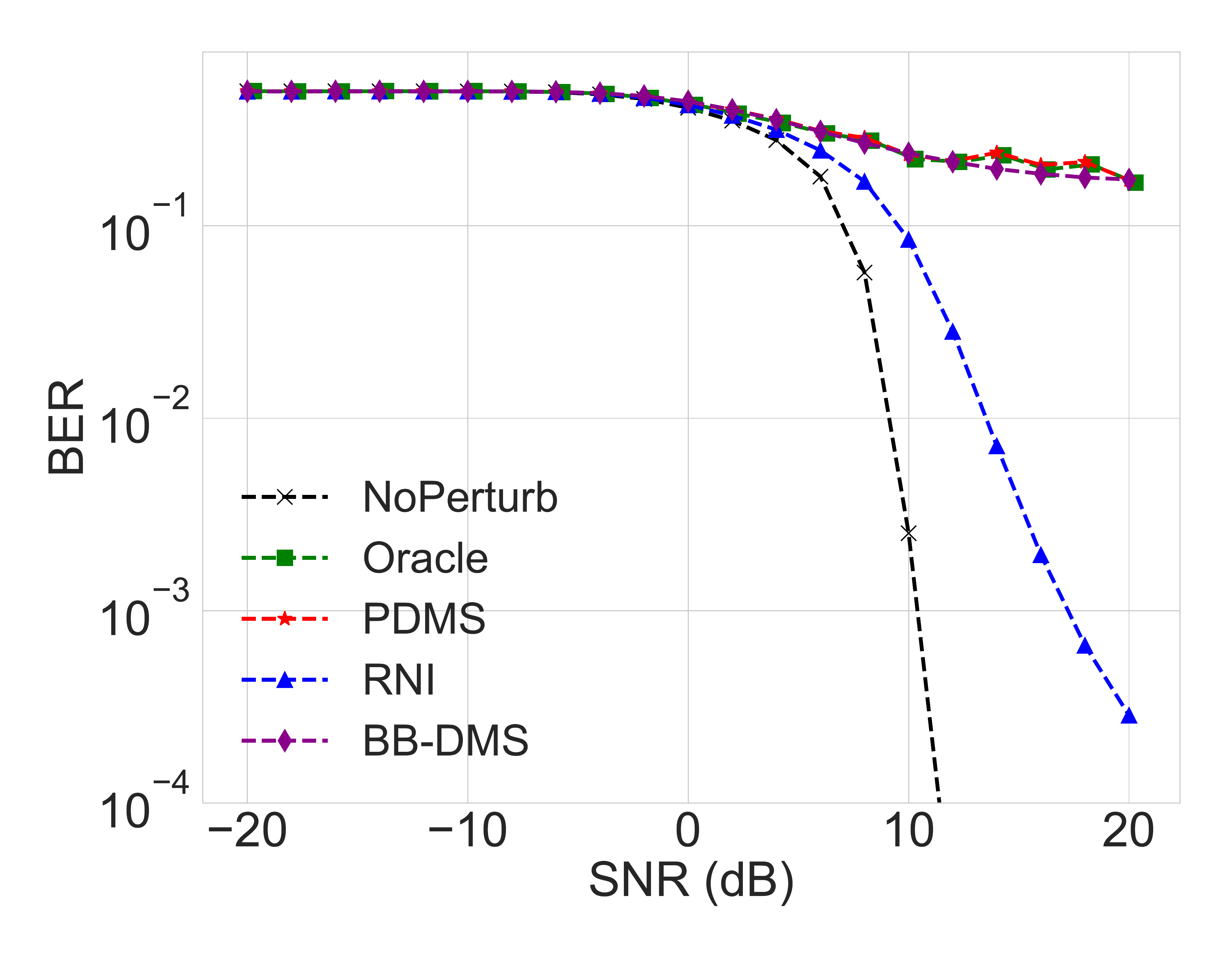}}
	\caption{BER vs SNR for \PSK8 and \QAM{64} modulated signal for different defensive modulation schemes.
	}
     \label{fig: ber_cr_2_by_3}
\end{figure}

For each modulation scheme, we generate data resulting in approximately 245000 I/Q channel symbols (note that for different modulation schemes this corresponds to different number of data bits), split into blocks of 128 I/Q symbols ($n=128$), as explained above. The last 300 blocks for each modulation scheme are reserved for testing the performance (tests are repeated 20 times), while we train a separate classifier for each SNR value based on the above data. As shown in  Fig.~\ref{fig: accu_cr_2_by_3} (see the curve with label \NoPerturb), for high SNR values the accuracy of the modulation classification is close to 90\%. 
As expected, the classification accuracy degrades as the SNR decreases (as the noise masks the signal), but even at -10 dB, the intruder can achieve a 40\% detection accuracy (as opposed to the 12.5\% accuracy a completely random detector would achieve).

In the experiments, we compare this performance with
\begin{itemize}
    \item our three defensive modulation schemes, \PDMS, \BDMS, and \BODMS, as well as \Oracle as baseline;
    \item adding uniform random noise of $L_2$-norm $\epsilon$ to a block, called \emph{random noise insertion} (\RNI), which is then normalized for power constraints;
    \item a black-box attack mechanism that does not use the classifier of the intruder, but calculates \PDMS against a classifier that has the same architecture as that of the intruder's but is trained separately (assuming no channel noise); we call this the \emph{Blackbox} \DMS (\BBDMS).
\end{itemize}

\indent All the above schemes, except for \RNI, are implemented using the projected (normalized) gradient descent (PGD) \cite{madry2017towards} method from the CleverHans Library \cite{papernot2018cleverhans-orig}, with 20 iterations, $\beta=0.2$ and $\epsilon=3$. $\epsilon=3$ results in significant reduction in modulation-classification accuracy without incurring too large BER at the intended receiver and has been determined by running experiments over different values of $\epsilon$. \RNI uses the same $\epsilon$. Note that a perturbation of this size accounts for about 7\% of the total energy of a block (which is 128 due to our normalization to the energy constraint). \Oracle serves as an upper bound on the achievable defensive performance given the parameters, while the role of \RNI is to analyze the effect of carefully crafted perturbations instead of selecting them randomly. \BBDMS explores the more practical situation where the exact classifier of the intruder is not known, but its training method and/or a similar classifier is available.

\subsection{Defensive modulation schemes with norm-bounded perturbations}
We first consider defensive modulation schemes with a bound on the $L_{2}$ norm of the applied perturbation. Fig.~\ref{fig: accu_cr_2_by_3} shows the modulation-classification accuracy for several methods. It can be seen that adding random noise (\RNI) helps very little compared to no defense at all (\NoPerturb). The basic defense mechanism \PDMS and its black-box version \BBDMS become effective from about -5 dB SNR, and; as expected; \PDMS outperforms \BBDMS. For smaller SNR values the classification accuracy is relatively small (the channel noise already makes classification hard), and only the oracle defense \Oracle gives noticeable improvement. As expected, the performance of \PDMS gets closer to its lower bound, \Oracle, as the SNR increases (note that the two methods coincide at the limit of infinite SNR).  The similar performance of \BBDMS and \PDMS for medium SNR values shows a similar transferability of adversarial attacks in our situation as was observed in other machine learning problems, such as in image classification \cite{papernot2017practical, papernot2016transferability},
although this effect deteriorates quickly as the SNR increases and \PDMS becomes more effective. Observe that the classification accuracy of \PDMS increases up to 0 dB SNR, when the channel noise during both the training phase and test phase is higher than the defensive perturbation and thus, channel noise is the main cause of the performance limitation of the intruder, while the accuracy decreases for higher SNR when the defensive perturbation is larger compared to the channel noise and the defense mechanisms start working.

\begin{figure*}[t]
\centering
      \subfloat[\CPFSK-original]{\includegraphics[width=0.25\textwidth]{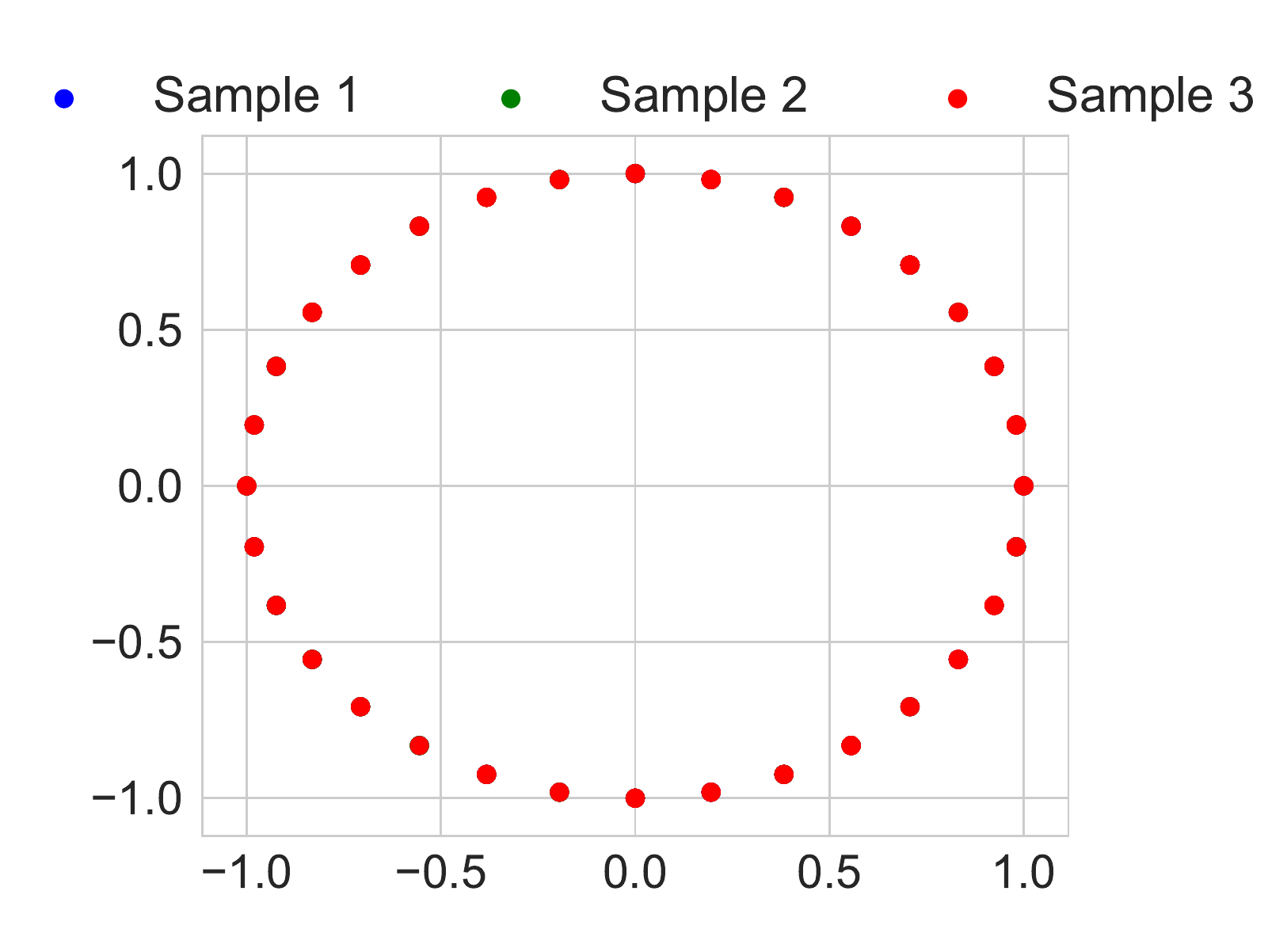}}
      \subfloat[\CPFSK-perturbed]{\includegraphics[width=0.25\textwidth]{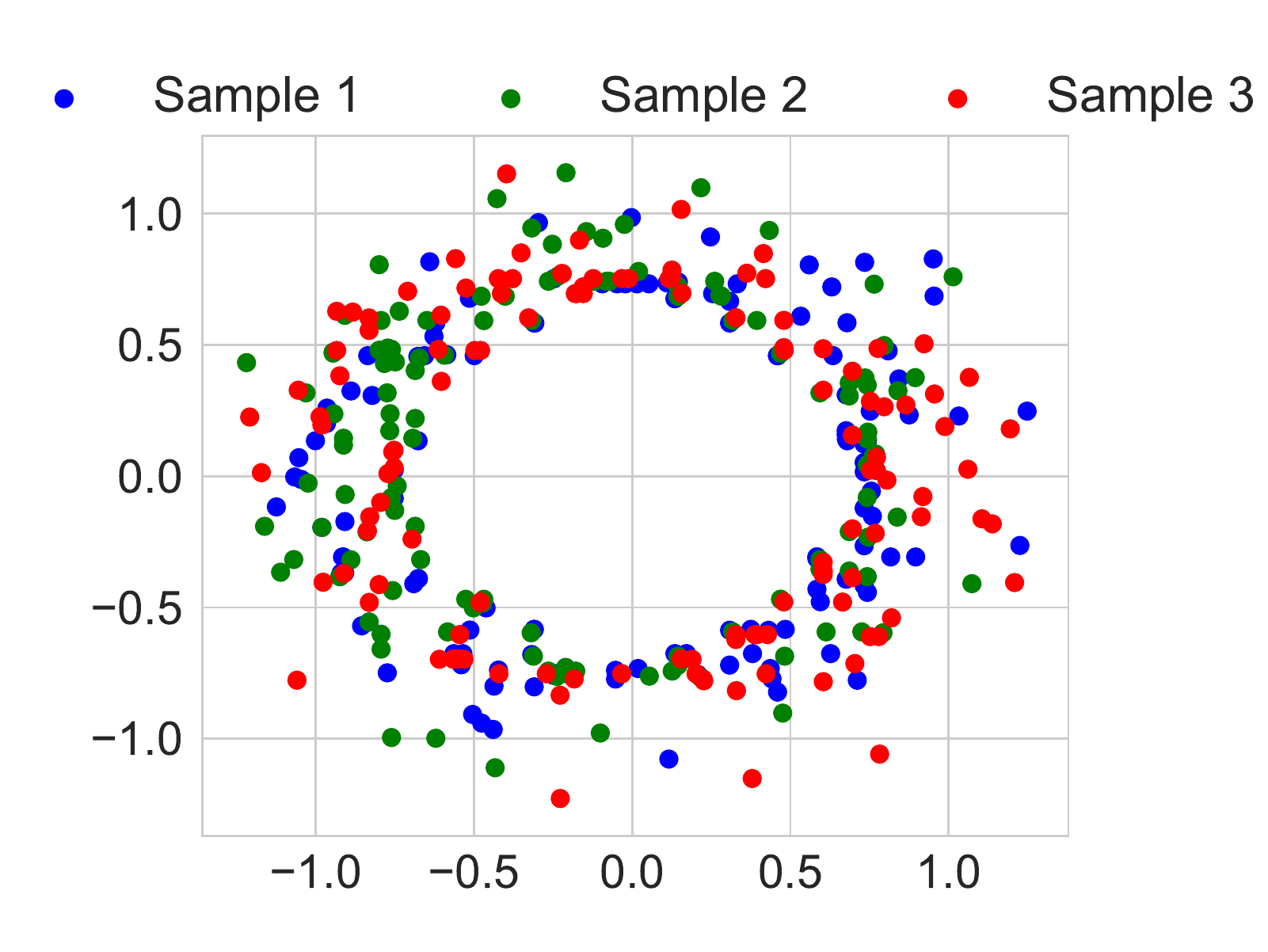}}
      \subfloat[\BPSK-original]{\includegraphics[width=0.25\textwidth]{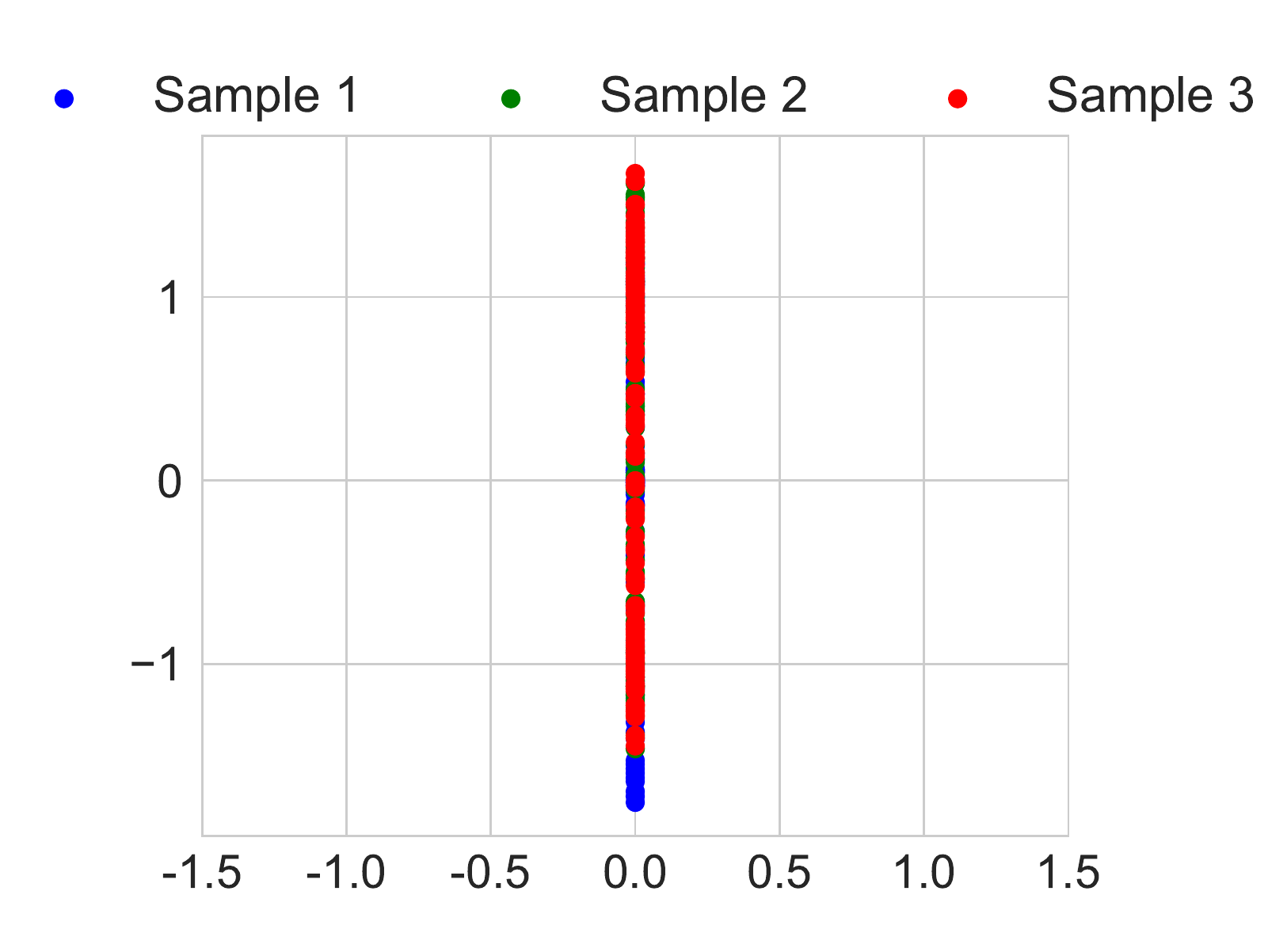}}
      \subfloat[\BPSK-perturbed]{\includegraphics[width=0.25\textwidth]{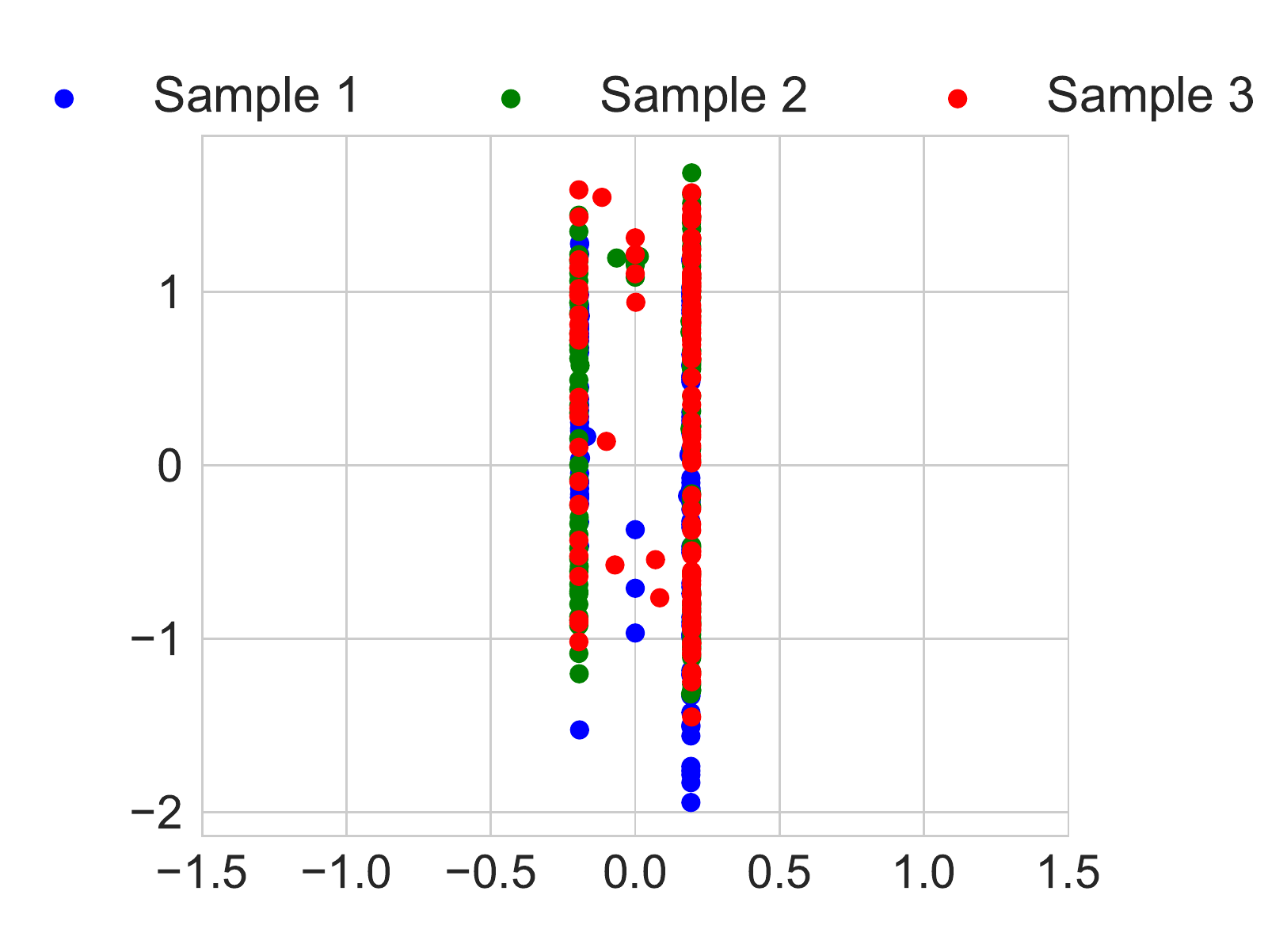}} \\
      \subfloat[\QAM{16}-original]{\includegraphics[width=0.25\textwidth]{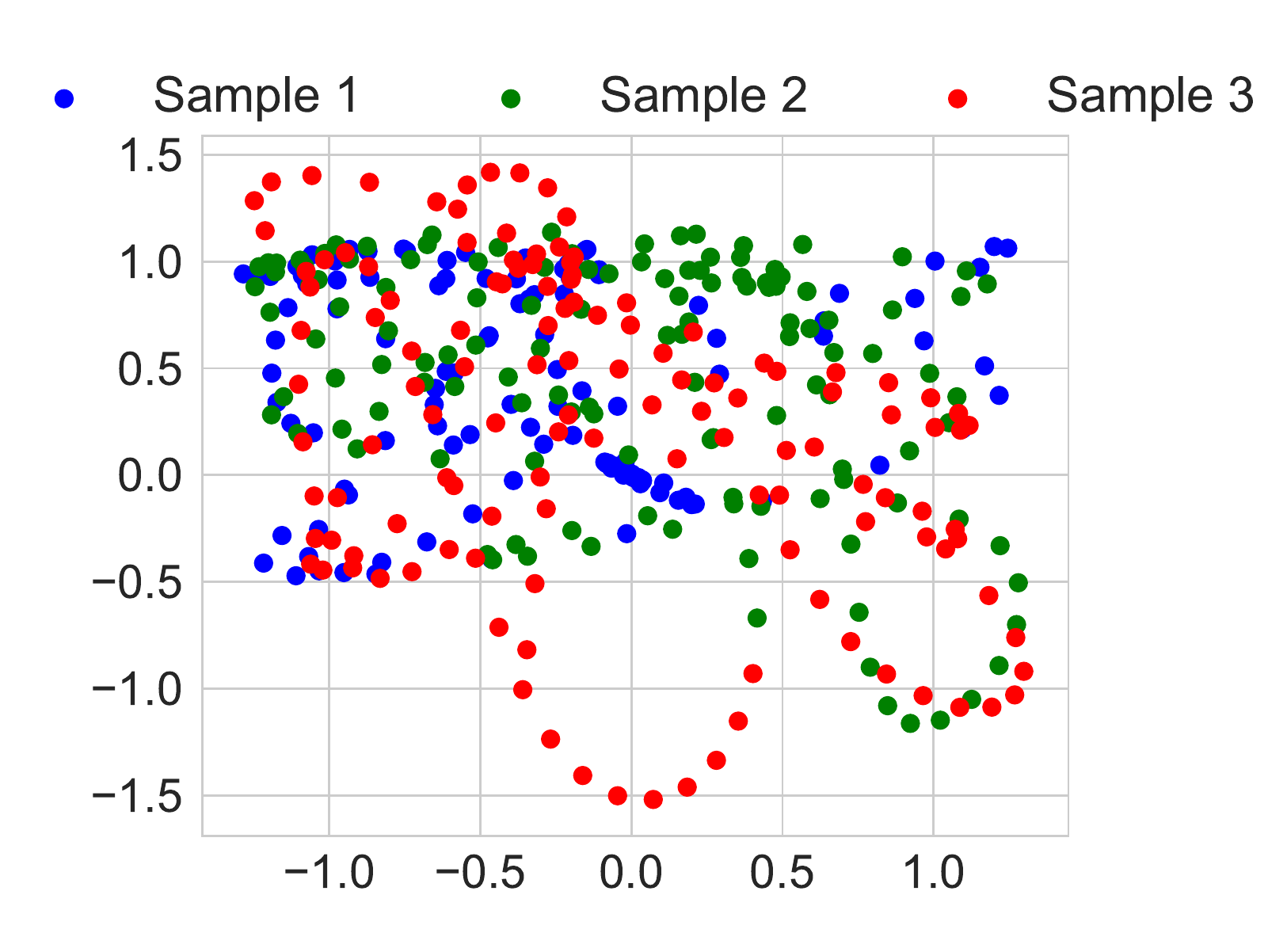}}
      \subfloat[\QAM{16}-perturbed]{\includegraphics[width=0.25\textwidth]{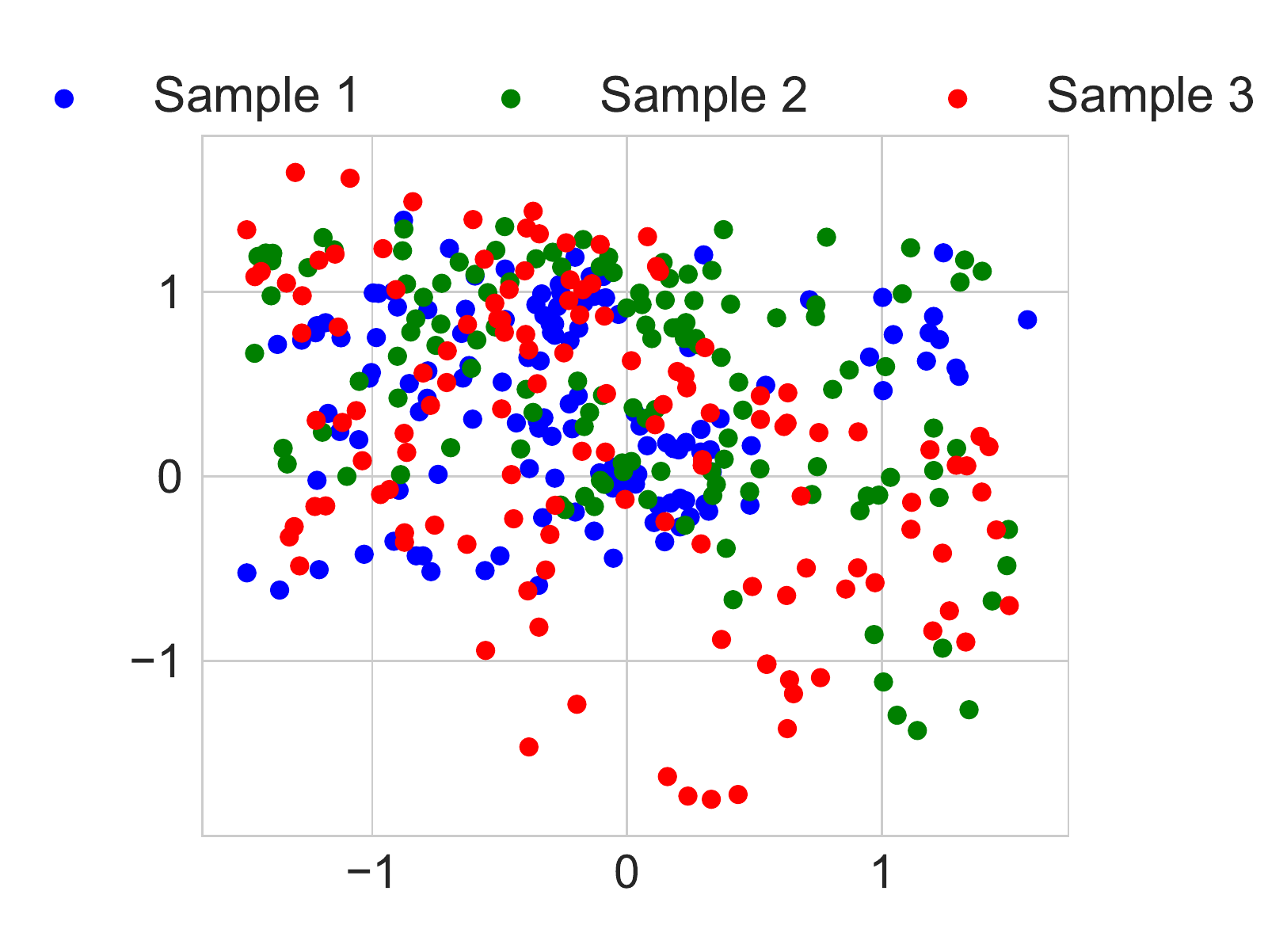}}
      \subfloat[\QAM{64}-original]{\includegraphics[width=0.25\textwidth]{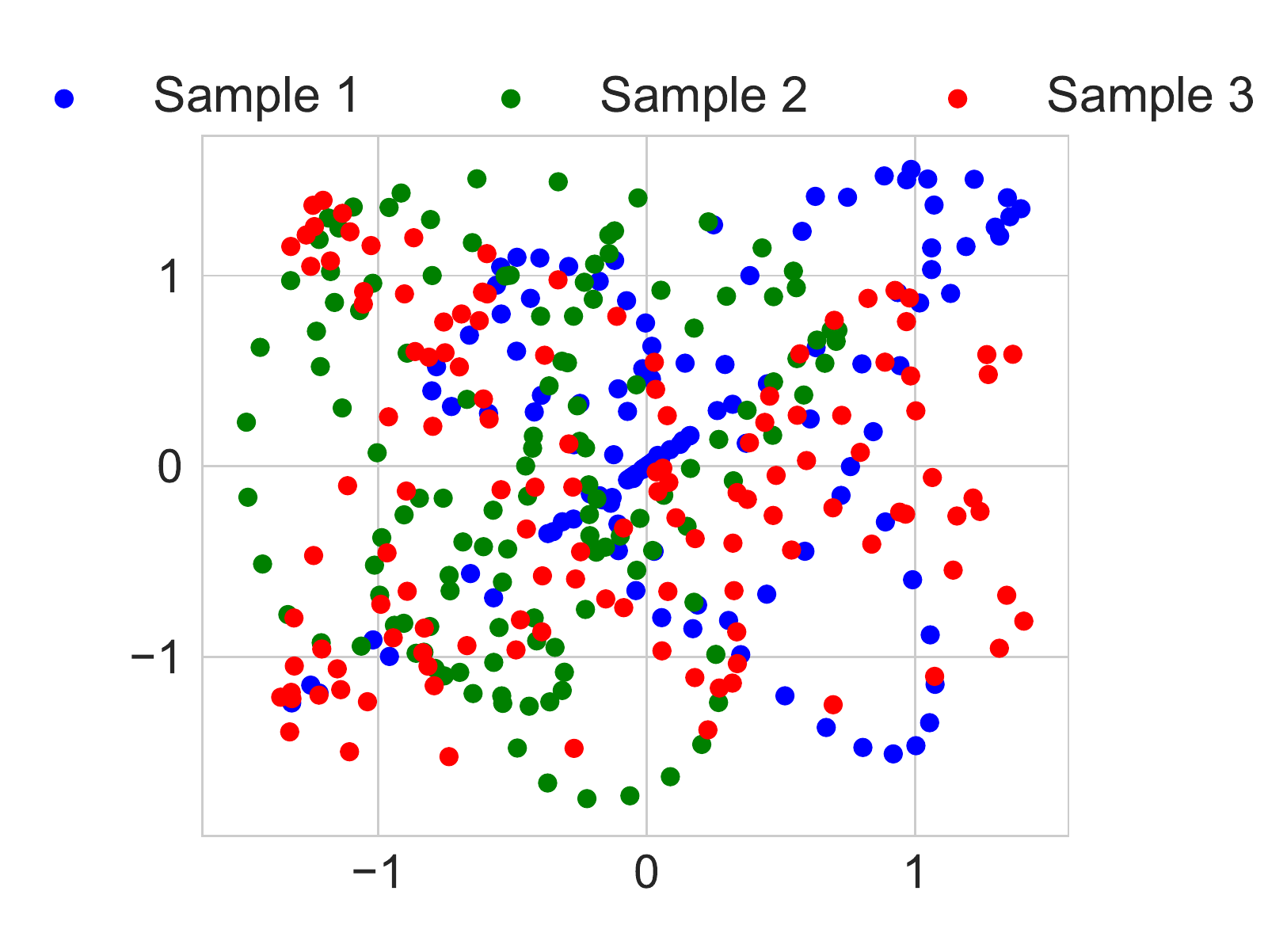}}
      \subfloat[\QAM{64}-perturbed]{\includegraphics[width=0.25\textwidth]{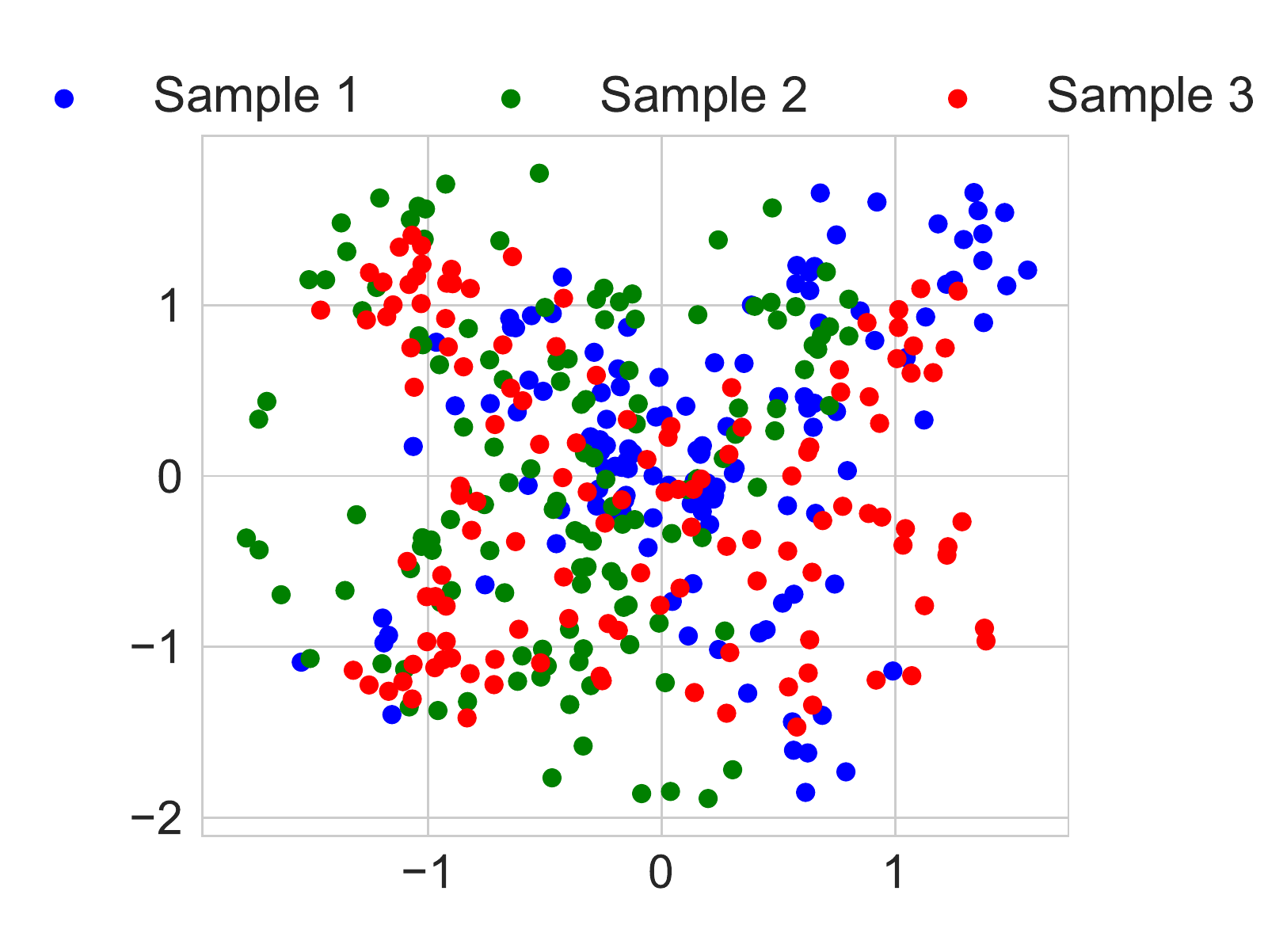}} \\
      \caption{Original constellation points and the perturbed channel input symbols with \PDMS for \CPFSK, \BPSK, \QAM{16} and \QAM{64} modulation schemes for first three inputs samples (3 $\times$ 128 channel symbols) to modulation classifier.}
      \label{fig: signal}
\end{figure*}

Table~\ref{tab: mod_schemes_accuracy} shows the modulation-classification accuracy for the individual modulation schemes at channel SNR of 20 dB. It can be seen that a defensive perturbation of the same norm $\epsilon$ affects different modulation schemes differently, where \CPFSK and \BPSK appear to be the most robust against defensive perturbations. Note that  \QAM{16} and \QAM{64} are very difficult to classify even without any perturbations, which is in line with the observation made in \cite{o2017introduction}. Modulated signals without any perturbation and \PDMS-modulated signals are presented in Fig.~\ref{fig: signal}, which shows that, even after perturbation, \CPFSK retains the modulated signal constellation and the perturbed \BPSK signals are still different from the output of any other modulation scheme. On the other hand, it becomes difficult to distinguish \QAM{16} and \QAM{64} signals.

\begin{table}[t]
\begin{center}
\begin{tabular}{l||c|c}
\hline
\textbf{Modulation} & \textbf{\NoPerturb} & \textbf{\PDMS} \\ \hline

\GFSK  & 1.0 & 0.02 \\
\hline
\CPFSK & 1.0 &  0.963 \\
\hline
\PSK8  & 0.986 & 0.0167 \\
\hline
\BPSK  & 1.0   & 0.76 \\
\hline
\QPSK  & 0.996 & 0.07 \\
\hline
\PAM4  & 1.0   & 0.096 \\
\hline
\QAM{16} & 0.48  & 0.376 \\
\hline
\QAM{64} & 0.526 & 0.376 \\
\hline
\end{tabular}
\end{center}
\caption{Classification accuracy of \PDMS for different modulation schemes with $\epsilon=3$ at SNR 20 dB.}
\label{tab: mod_schemes_accuracy}
\end{table}

The reduced classification accuracy of the intruder for \PDMS and \BBDMS are countered by the increased BER at the receiver. To illustrate this effect, Fig.~\ref{fig: ber_cr_2_by_3} shows the BER for \PSK8 and \QAM{64}; the other modulation schemes, except for \QAM{16}, show similar relative behavior to \PSK8, but with the error dropping sharply for medium SNR values, with a few dB difference among different modulation schemes (up to about 5 dB for \PSK8). On the other hand, the price of using any defense mechanism on \QAM{64} is severe, resulting in a significantly higher BER in the high SNR regime; \QAM{16} behaves similarly with somewhat smaller BER values. For the \Oracle defensive scheme, we directly feed the perturbed signal to the decoder to calculate the BER, which is lower than the BER at the decoder when the \PDMS and \BBDMS defensive schemes are employed for \PSK8, while these BERs are essentially the same for \QAM{64} .

\begin{figure}[t]
\centering
   \subfloat[Modulation-classification accuracy vs SPR.  \label{subfig-1:accu_spr}]{%
       \includegraphics[width=\graphwidth]{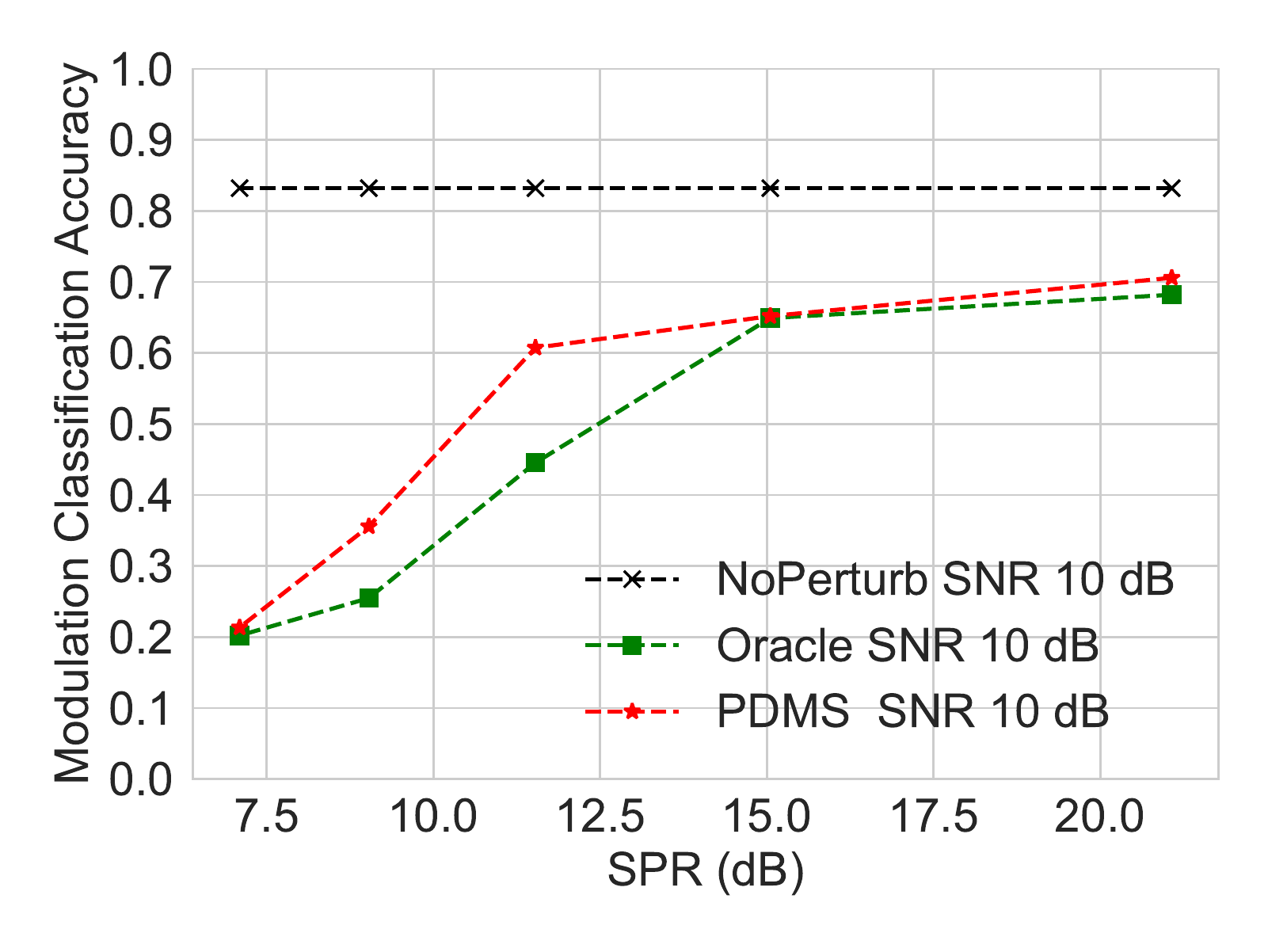}
     }
     \hfill
     \subfloat[BER vs SPR. \label{subfig-2:qam64_spr}]{%
       \includegraphics[width=\graphwidth]{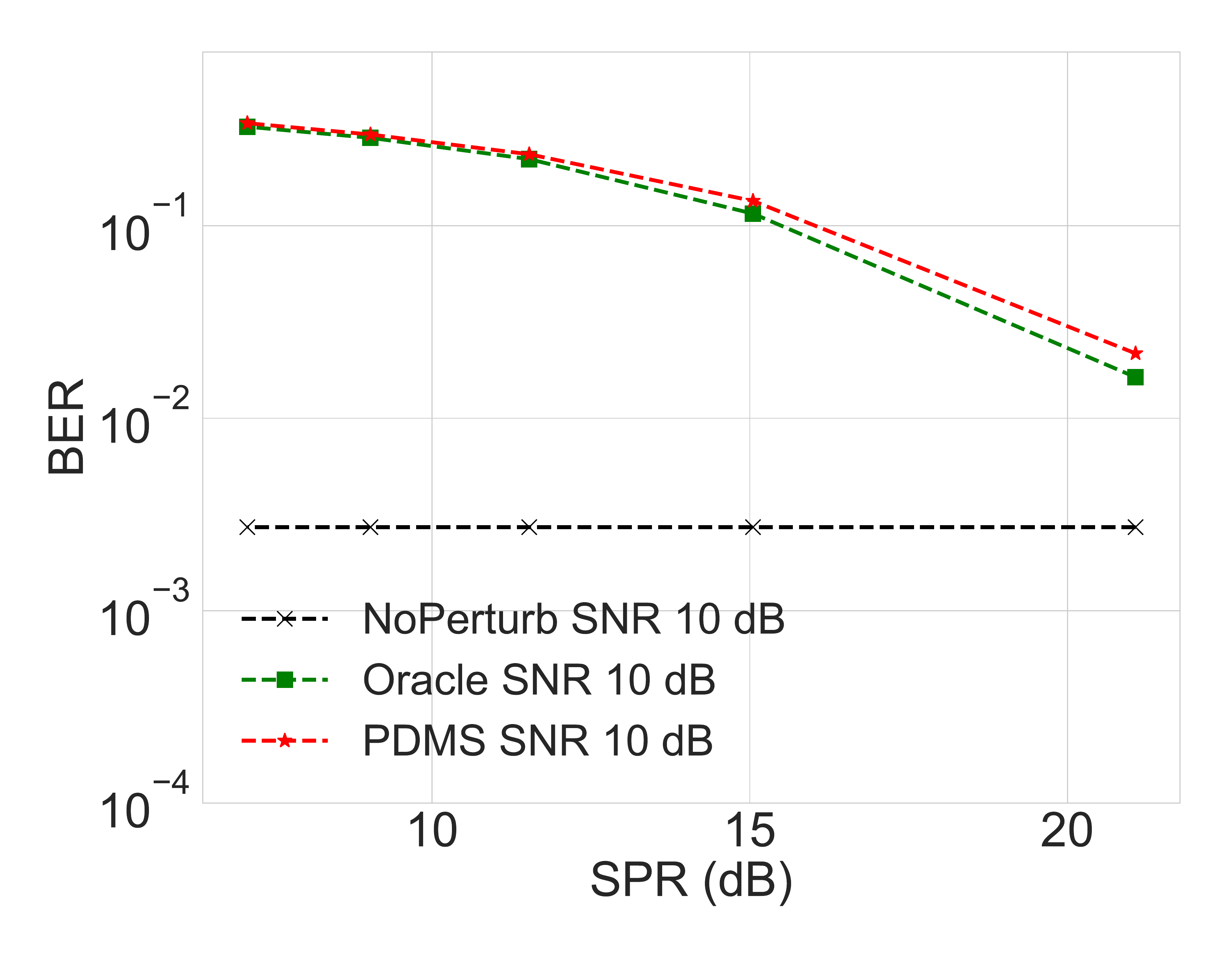}}
	\caption{Effect of signal-to-perturbation ratio (SPR) on the modulation-classification accuracy and the BER (\QAM{64}).}
     \label{fig: spr_cr_2_by_3}
\end{figure}

This negative effect on the BER can be suppressed if the perturbation size is decreased, which, at the same time, results in increased detection accuracy. This is shown in Fig.~\ref{fig: spr_cr_2_by_3} as a function of the signal-to-perturbation ratio $SPR \triangleq n/\|\bdelta\|_2^2$ (recall $n=128$, and $SPR\approx 11.5 $dB corresponds to $\epsilon=3$).
In every case, \PDMS trades off increased BER for reduced detection accuracy compared to the case when no defense mechanism is applied. Also, increasing the number of iterations used in the defensive schemes to compute the perturbations has limited impact on modulation-classification accuracy and BER as the total perturbation is limited to have $L_2$-norm $\epsilon$.

\begin{figure}[t]
\centering
      \includegraphics[width=\graphwidth]{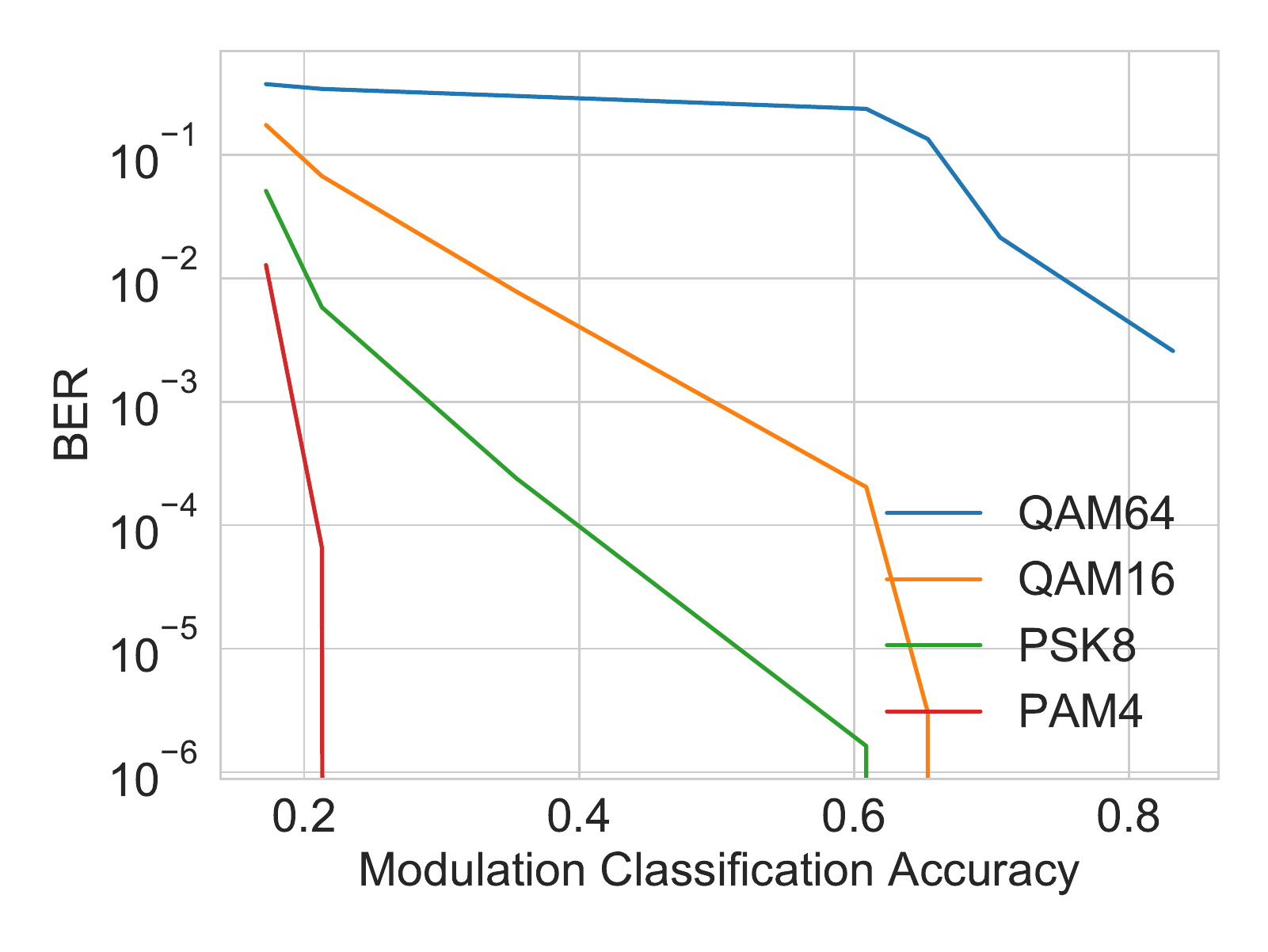}
	\caption{Trade-off between the modulation-classification accuracy  and the BER for \PDMS with code rate $2/3$, where the $L_2$ norm of the perturbations is limited by $\epsilon \in [1,6]$. The accuracy is averaged over all modulation schemes while the BER is shown for each modulation scheme separately. \BPSK, \QPSK, \GFSK and \CPFSK have zero error rate for these perturbations.}
     \label{fig: Accu_vs_ber}
\end{figure}

Fig.~\ref{fig: Accu_vs_ber} shows the trade-off between the average modulation-detection accuracy of the intruder and the BER for the individual modulation schemes for an intruder DNN trained at an SNR of 10 dB (i.e., the training samples are generated with this channel SNR) when the maximum perturbation norm $\epsilon$ of \PDMS takes values in the range $[1,6]$ (smaller $\epsilon$ values correspond to points with smaller BER and larger classification accuracy on each curve). It can be seen that an effective perturbation that results in a reduction in the modulation-classification accuracy also causes an increase in the BER. The trade-off between the two is different for different modulation schemes for the same perturbation constraint $\epsilon$  (note that the reported classification accuracy is an average computed over all modulation schemes) . It can be seen that an increase in $\epsilon$ needed to reduce the average modulation-classification accuracy results in large BER for \QAM{16} and \QAM{64}. Note that in our experiments \BPSK, \QPSK, \GFSK and \CPFSK have zero error rate for this $\epsilon$ range, hence they are not included in the figure.

\subsection{BER-aware defense schemes}

A more systematic way of improving the BER is to use our BER-aware modulation schemes \BDMS and \BODMS.
In the numerical experiments, due to the large computational overhead of calculating the SPSA gradient estimates in \eqref{eq:SPSA} (with $K=400$), we only used 400 signal blocks to measure the test performance (instead of the $300\times20=6000$ blocks used previously).
Also, to keep the required computation feasible, in \eqref{eq:SPSA} we used error rates calculated over 100 signal blocks (that is, over 12800 perturbed channel input symbols simultaneously). This approximation allowed us to run Viterbi decoding once for every hundred blocks, instead of running it from the beginning for every block, causing a substantial reduction in computational complexity. The approximate gradient of $e$ computed this way was then used to calculate one step of the optimization (i.e., the next candidate perturbation) for each of the 100 blocks simultaneously. The drawback of this approximation is twofold: (i) instead of taking the gradient for a single perturbation, for each perturbation the error gradient is computed as an average coming from perturbing each of the 100 blocks simultaneously (this affects negatively the accuracy of the optimization); (ii) the applied method introduces delays in the transmission as it assumes that all signal blocks perturbed together are available at the transmitter at the same time (this gives some optimistic bias to the optimization compared to non-delayed real-time encoding).
Nevertheless, we believe that the negative effects are stronger here, and the performance of our modulation schemes (\BDMS and \BODMS) could be improved if the BER of the individual signal blocks were used for gradient estimation in SPSA.

\begin{figure}[t]
\centering

  \includegraphics[width=\graphwidth]{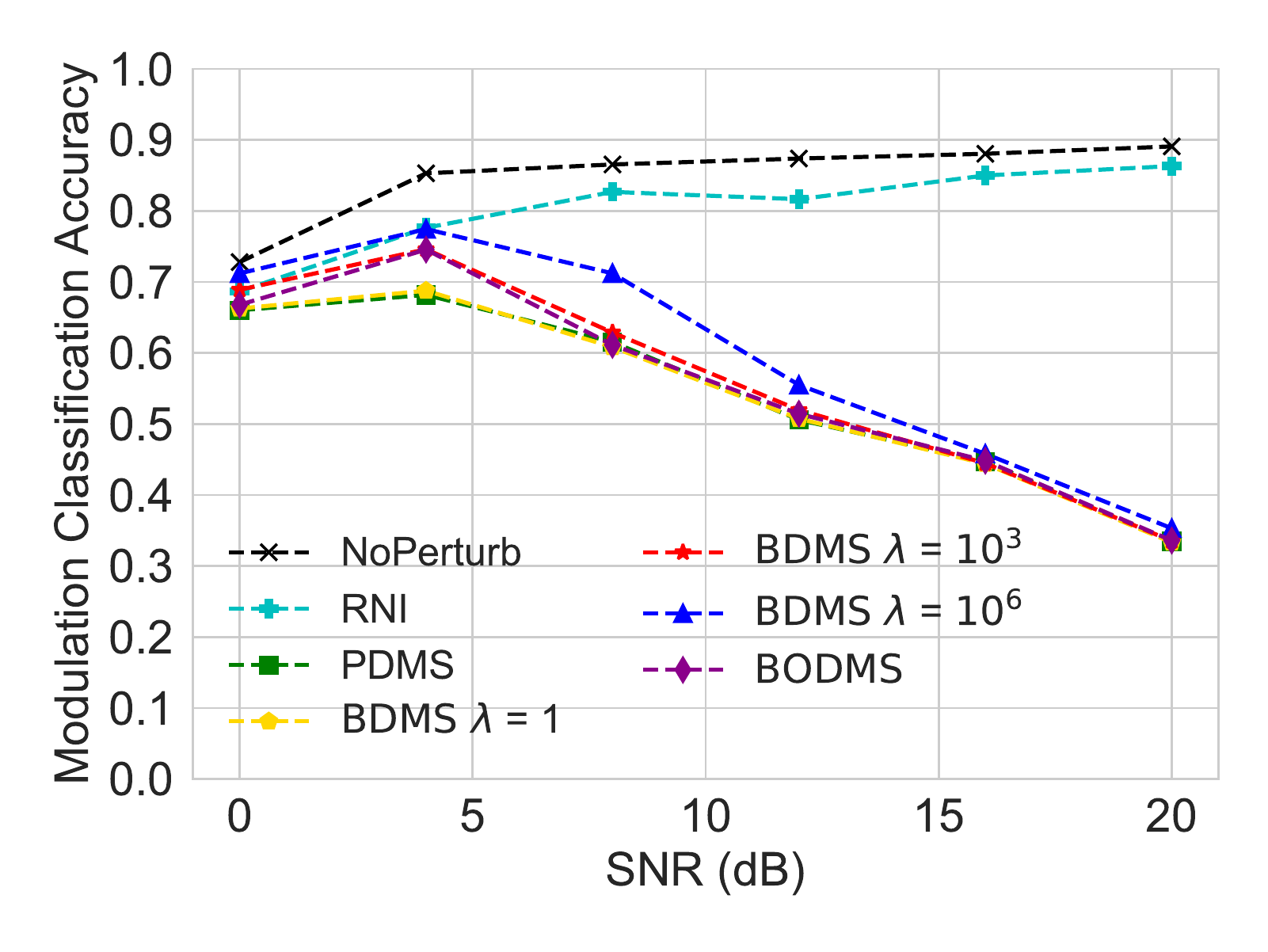}
  \caption{Modulation-classification accuracy of BER-aware defense mechanisms ($\epsilon$ = 3).}
  \label{fig: accuracy_spsa_dnn}

    \subfloat[ \vspace{-0.2cm}\PSK8 \label{subfig-1:spsa_psk8_dnn}]{%
       \includegraphics[width=\graphwidth]{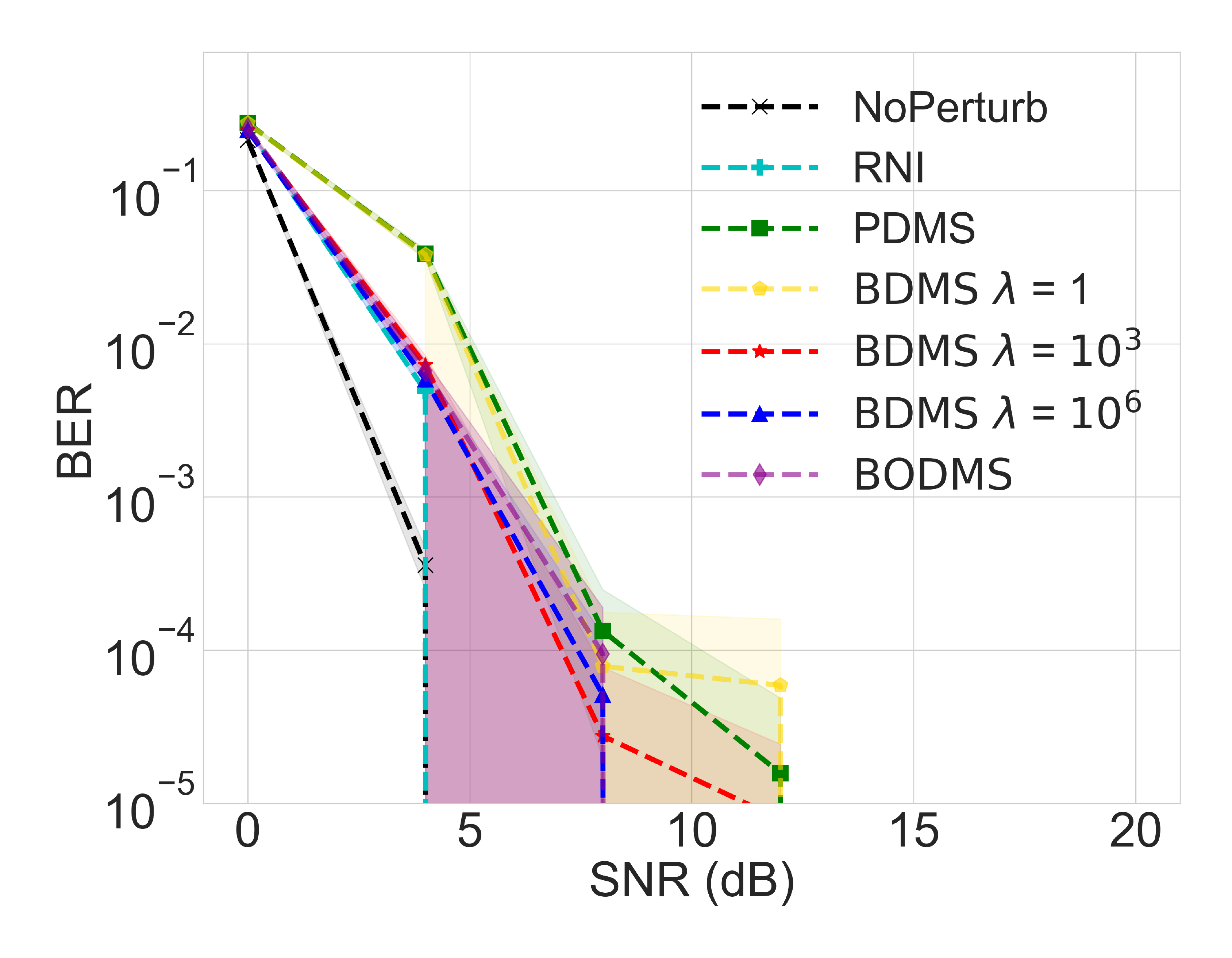}}
     \hfill
     \subfloat[\QAM{64} \label{subfig-2:spsa_qam64_dnn}]{%
       \includegraphics[width=\graphwidth]{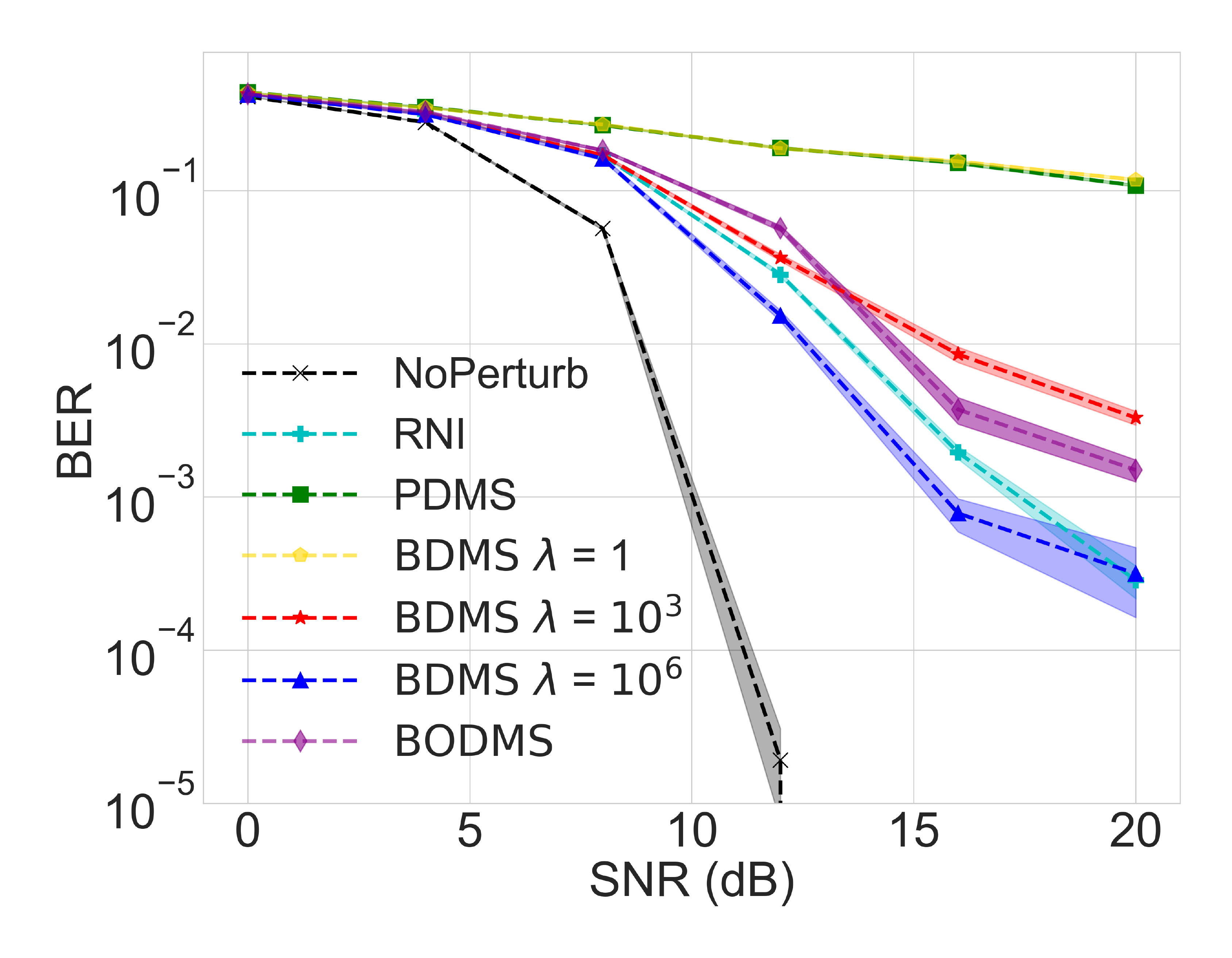}}
	\caption{BER for \PSK8 and \QAM{64} for BER-aware modulation schemes ($\epsilon$ = 3). In figure (a) the BER for \RNI is 0 beyond 5 dB, and all BER values are 0 when the SNR is larger than 12 dB.}
     \label{fig: ber_spsa_dnn}
     \vspace{-0.3cm}
\end{figure}

Fig.~\ref{fig: accuracy_spsa_dnn} and Fig.~\ref{fig: ber_spsa_dnn} show, respectively, the modulation-classification accuracy and the BER for \BDMS and \BODMS, also compared to \PDMS, \RNI and \NoPerturb, against a DNN-based intruder, which is trained with channel input symbols at specific SNR values. The performance of \BDMS is presented for three different values of $\lambda$, namely $1, 10^{3}, 10^{6} $. As before, the BER is shown for \PSK8 and \QAM{64}, as again \QAM{64} is the modulation scheme most affected by our perturbations, and except for \QAM{16} (which is similar to \QAM{64}), and the error rate for the other modulation schemes is similar to (in fact smaller than) that of \PSK8 and is very small under any defense mechanisms at high SNR values.

It can be seen that at high SNR (at least 12 dB), all defensive schemes achieve roughly the same classification accuracy, while \BODMS and \BDMS for large $\lambda$ provide significant improvement in the BER (shown for \PSK8 and \QAM{64}). Note, however, that the errors are still significantly higher than for the standard \QAM{64} modulation with no perturbation.

For larger $\lambda$ values, the BER of \BDMS for \QAM{64} is smaller than or approximately the same as for \RNI, which adds uniform random noise of the same perturbation size, while it significantly outperforms \RNI in classification accuracy (for \PSK8, both \RNI and \BDMS achieve low BER, although it can be much smaller for \RNI). Note that \BODMS approaches the performance of \BDMS with a large $\lambda$ ($10^3-10^6$), without the need to tune the hyperparameter $\lambda$, and these methods provide a good compromise between the effectiveness of the defense and the increase in the BER.

\begin{figure}[t]
\centering
   \subfloat[Modulation-classification accuracy  \label{subfig-1:accuracy_spsa_tree}]{%
       \includegraphics[width=\graphwidth]{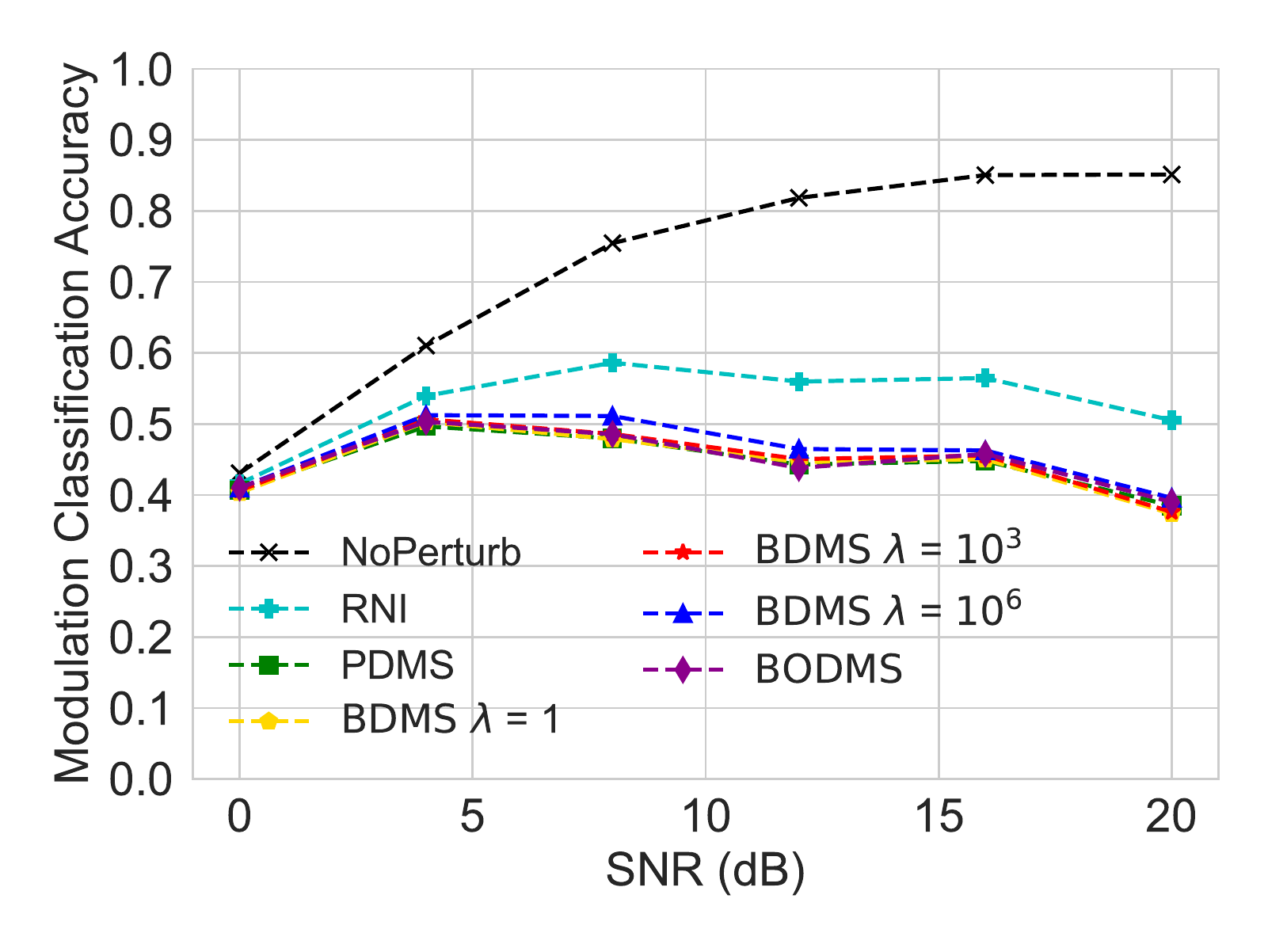}
     }
     \hfill
     \subfloat[BER \label{subfig-2:spsa_qam64_tree}]{%
       \includegraphics[width=\graphwidth]{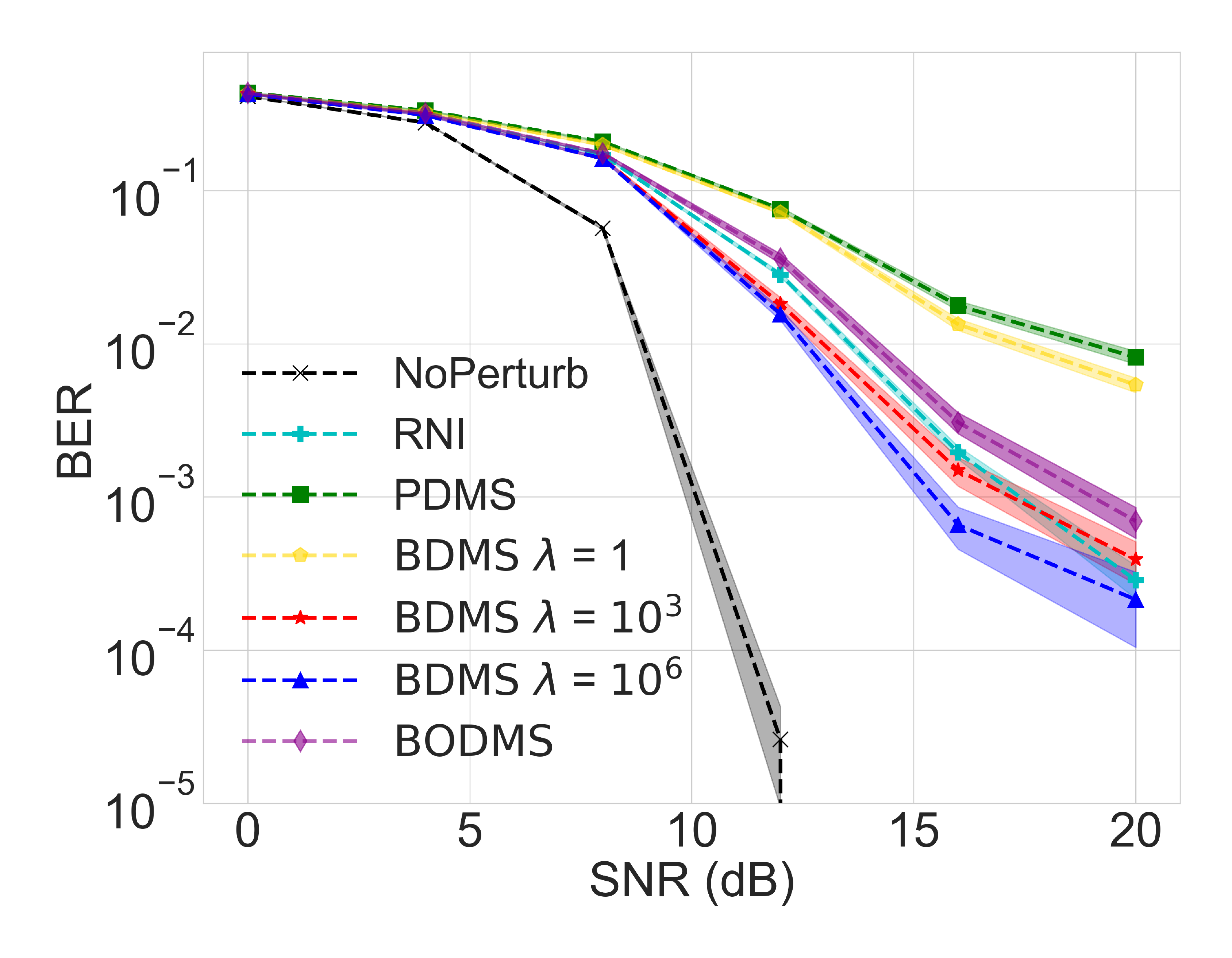}}
	\caption{Modulation-classification accuracy of tree-based intruder and BER (QAM{64}) for BER-aware modulation schemes ($\epsilon$ = 3).}
     \label{fig: spsa_tree}
     \vspace{-0.3cm}
\end{figure}

In addition to DNN-based detectors at the intruder, we also examine defense against one of the best standard modulation detection schemes in the literature, a multi-class decision tree trained with expert features obtained from \cite{Rosti98statisticalmethods, abdelmutalab2016automatic}.
Fig.~\ref{fig: spsa_tree} shows the modulation-classification accuracy and the BER achieved by employing various defense mechanisms against this intruder. It can be seen that the BER achieved against the tree-based classifier is approximately the same as the one achieved against the DNN-based classifier with \BDMS and \BODMS, while the accuracy of the DNN-based classifier is consistently higher, except for some high SNR values, when they are approximately the same. This demonstrates that our observations and conclusions also apply to intruders employing other types of detection mechanisms.

\subsection{Robustness of the intruder's classifier}\label{sec: robustness}

\begin{figure}[t]
     \centering
      \subfloat[\vspace{-0.15cm}No defense (\NoPerturb) \label{subfig-1:accu_non_adv}]{%
        \includegraphics[width=\graphwidth]{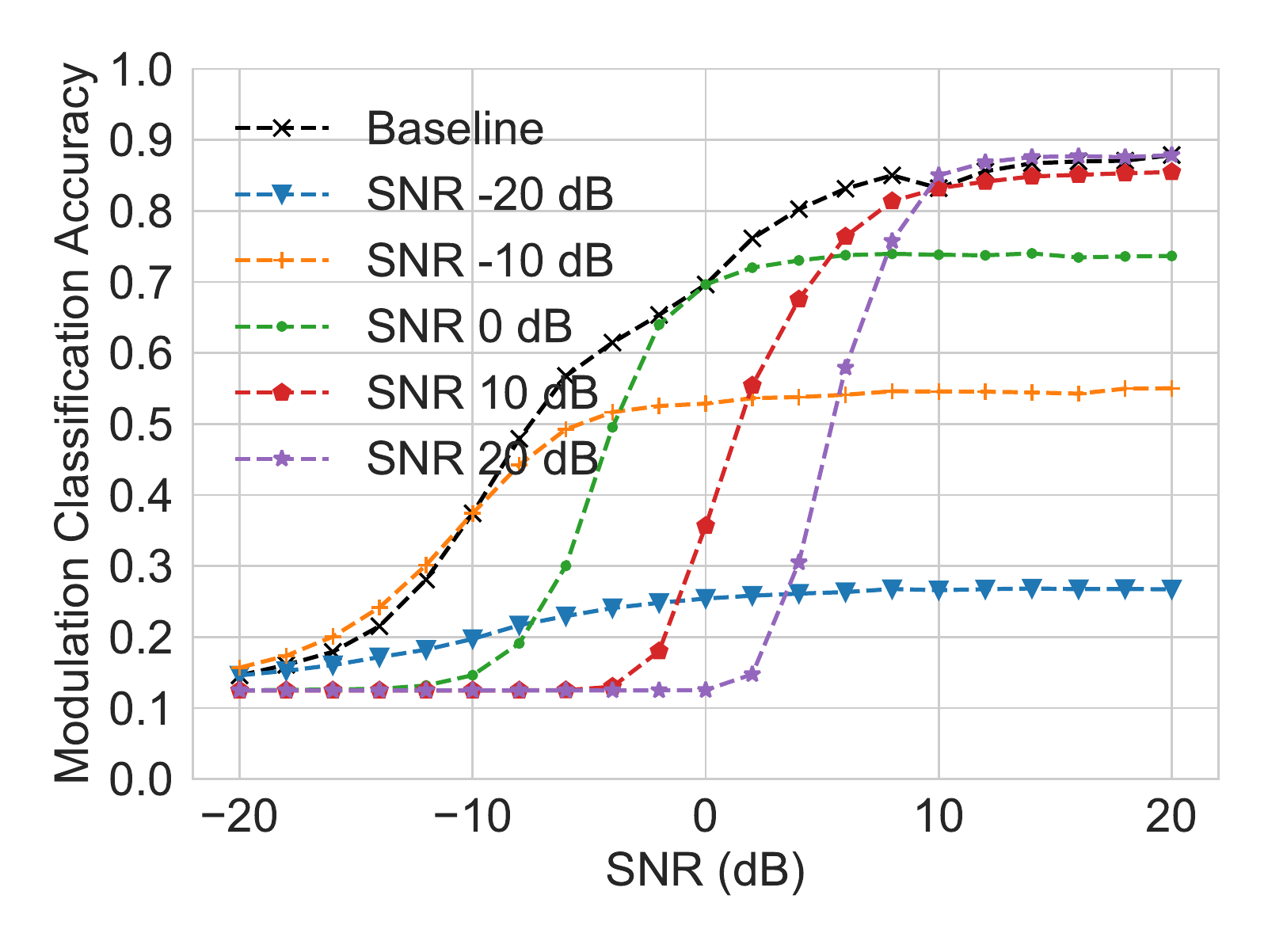}
      }
      \hfill

      \subfloat[\vspace{-0.15cm}\RNI \label{subfig-2:accu_tx_uniform}]{%
        \includegraphics[width=\graphwidth]{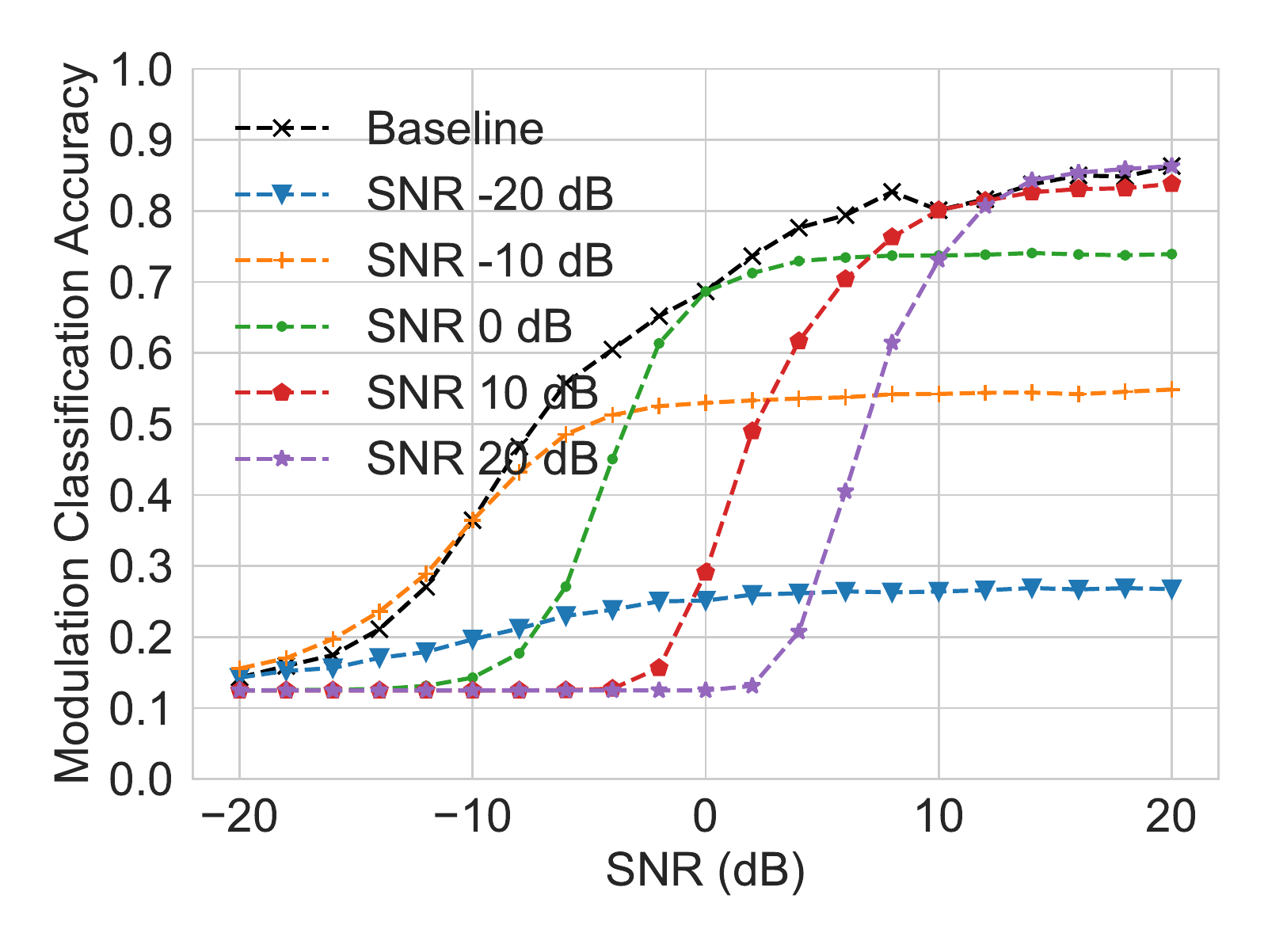}
      }

      \hfill

      \subfloat[
      \PDMS\label{subfig-4:accu_tx_adv}]{%
        \includegraphics[width=\graphwidth]{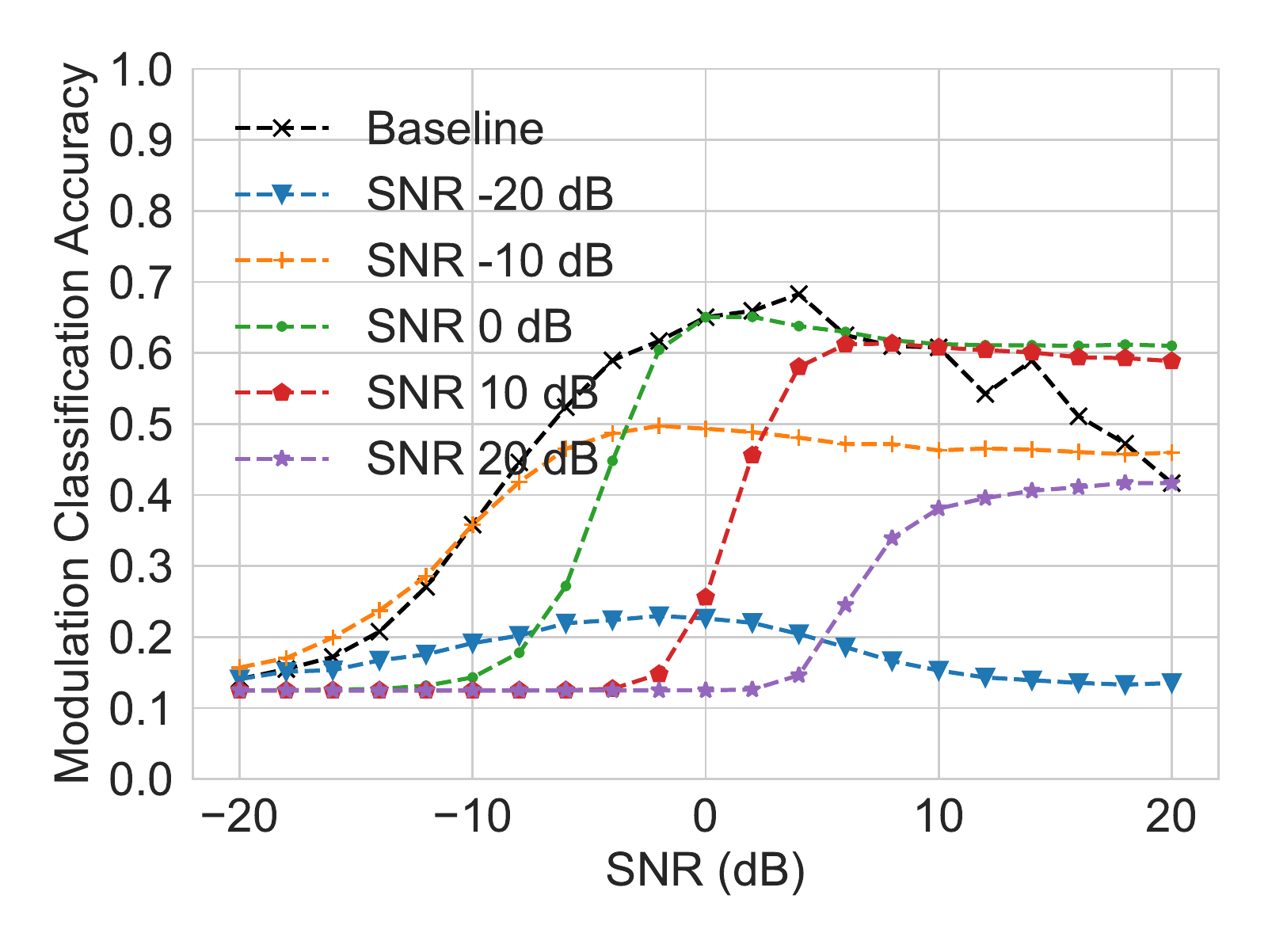}
      }
      \caption{Modulation-classification accuracy of the intruder as a function of the test channel SNR for total perturbation $\epsilon$  = 3.0}
      \label{fig: accu_different_attacks}
      \vspace{-0.3cm}
    \end{figure}

In the previous sections we assumed that the intruder knows the SNR of its received signals perfectly and trains its classifier for this SNR value. Although this may not be possible in practice due to estimation errors or variations in channel quality, assuming more accurate information at the intruder should allow us to design stronger defense mechanisms. In this subsection, we study the robustness of the intruder's detection network against errors in its SNR estimate; that is, we study its modulation-detection accuracy when it is trained for a specific channel SNR, but tested at different SNR values. We show in Fig.~\ref{fig: accu_different_attacks} the results for three cases: (a) when no defense mechanism is applied (i.e., \NoPerturb); (b) when uniform noise is added (\RNI); and
(c) when our perturbation-based defense \PDMS is applied. In each figure, we plot the detection accuracy with respect to the test channel SNR when the intruder is trained at five different SNR values. \Baseline represents the case in which the test channel SNR matches the training SNR.

We can observe in Fig.~\ref{subfig-1:accu_non_adv} that the intruder network trained at channel SNR -20 dB is unable to learn any effective classifier for higher SNR values. As the channel SNR at the time of training increases, its performance improves for a larger range of test SNR values as evident from the plots for SNR -10 dB and 0 dB, but, as one would expect, the accuracy achieved is below the peak accuracy values in the \Baseline curve. On the other hand, networks trained with high SNR values of 10 dB and 20 dB achieve higher accuracy, close to peak accuracy values in the \Baseline curve, but tend to breakdown when SNR goes below a certain value (2 dB and 6 dB for intruder networks trained at 10 dB and 20 dB, respectively). It is due to the fact the DNNs learn the classifier function (decision boundaries) from the training data, and for those trained at high channel SNR, signals with higher noise may lie across decision boundaries learned from less noisy training data, and are wrongly classified.

Note that perturbations in Figs.~\ref{subfig-2:accu_tx_uniform} and~\ref{subfig-4:accu_tx_adv} are generated with total $L_{2}$-norm $\epsilon=3.0$ for each trained network and at each SNR value. It can be seen from Fig.~\ref{subfig-2:accu_tx_uniform} that adding random perturbations does not reduce the modulation-classification accuracy, yielding similar performance to the case when no defense mechanism is applied (Fig.~\ref{subfig-1:accu_non_adv}).

When \PDMS is employed, if the network is trained for a low SNR value, then test data with lower noise level (higher SNR) will lie at a larger distance from the decision boundaries (learned from noisy data), since the decision boundaries are already accounting for a very high noise level. Therefore, in this case the total perturbation $\epsilon$ may not be enough to move the signal to the wrong side of a learned decision boundary of the intruder, resulting in a higher accuracy. On the other hand, when the network is trained for a higher SNR value than the test channel, there is not much variation in the training data due to the absence of noise, and an attacks with even limited perturbation is enough to move the data point to the other side of the learned decision boundary, changing the class label.

In case of \PDMS, the intruder networks trained at low channel SNR values of 0 dB and 10 dB are more robust against  \PDMS as the learned decision boundary accounts for larger channel noise during training of the intruder NN and perturbation norm $\epsilon$ is too small in comparison to channel noise at smaller test SNR values to move the perturbed signal across it. Once the defensive perturbation $\epsilon$ becomes comparable in magnitude to the test data channel SNR then both intruder networks show similar performance for test SNR ($\geq$ 5 DB). In the case of an intruder network trained at SNR 20 dB, perturbation $\epsilon$ is large compared to channel noise for higher test SNR values, and thus, results in small modulation-detection accuracy. Also, since the signal is perturbed before transmission, these defensive perturbations are partially masked by the channel noise. This effect of the channel noise is prominent in accuracy curves, though \PDMS perturbations significantly reduce the detection accuracy as evident in Fig.~\ref{subfig-4:accu_tx_adv}.

\begin{figure}[t]
\centering
   \subfloat[Without curriculum training.  \label{subfig-1:accuracy_mix_dnn_complete_noncur}]{%
       \includegraphics[width=\graphwidth]{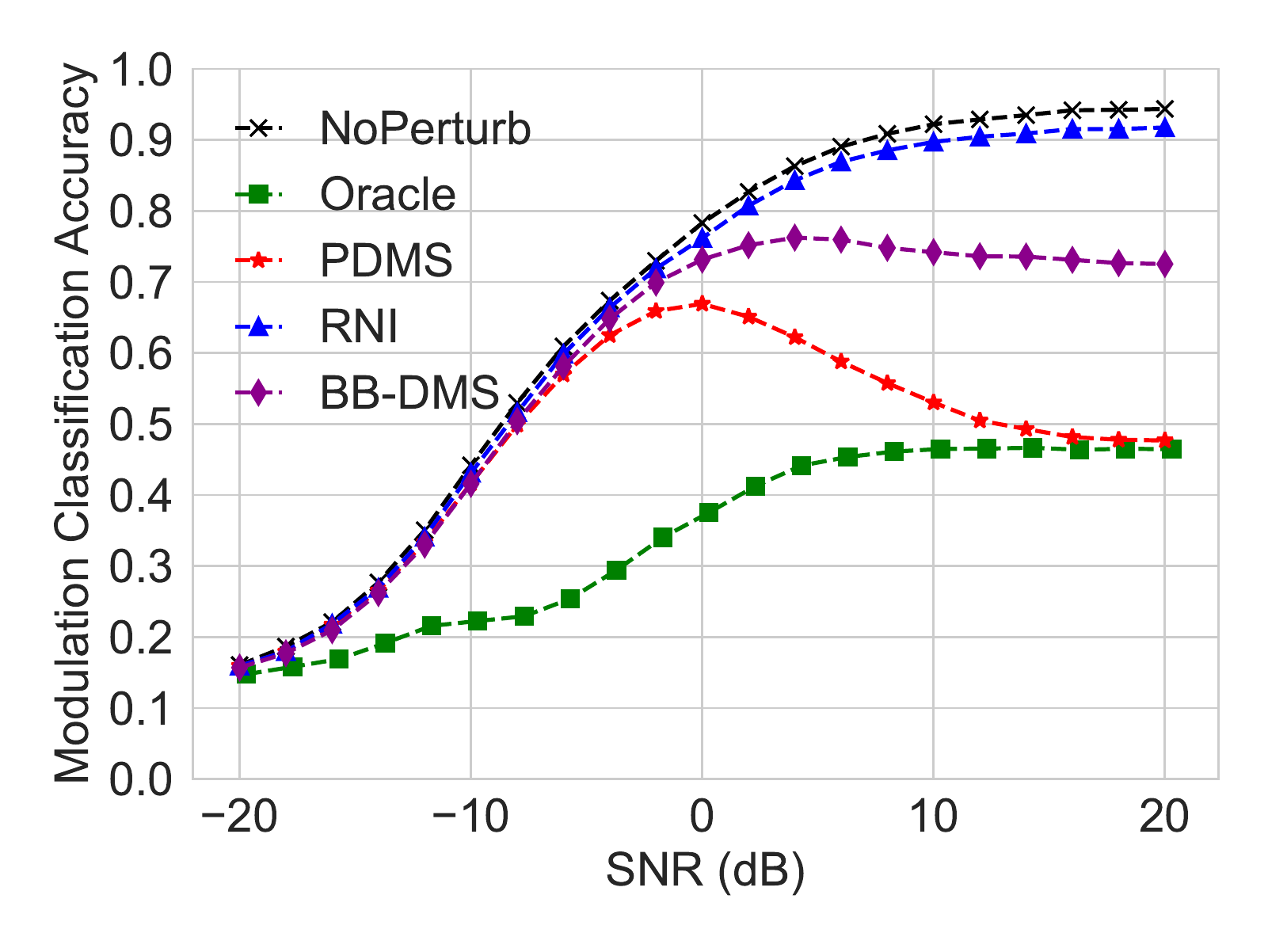}
     }
     \hfill
       \subfloat[With curriculum training. \label{subfig-1:accuracy_mix_dnn_complete_cur}]{%
       \includegraphics[width=\graphwidth]{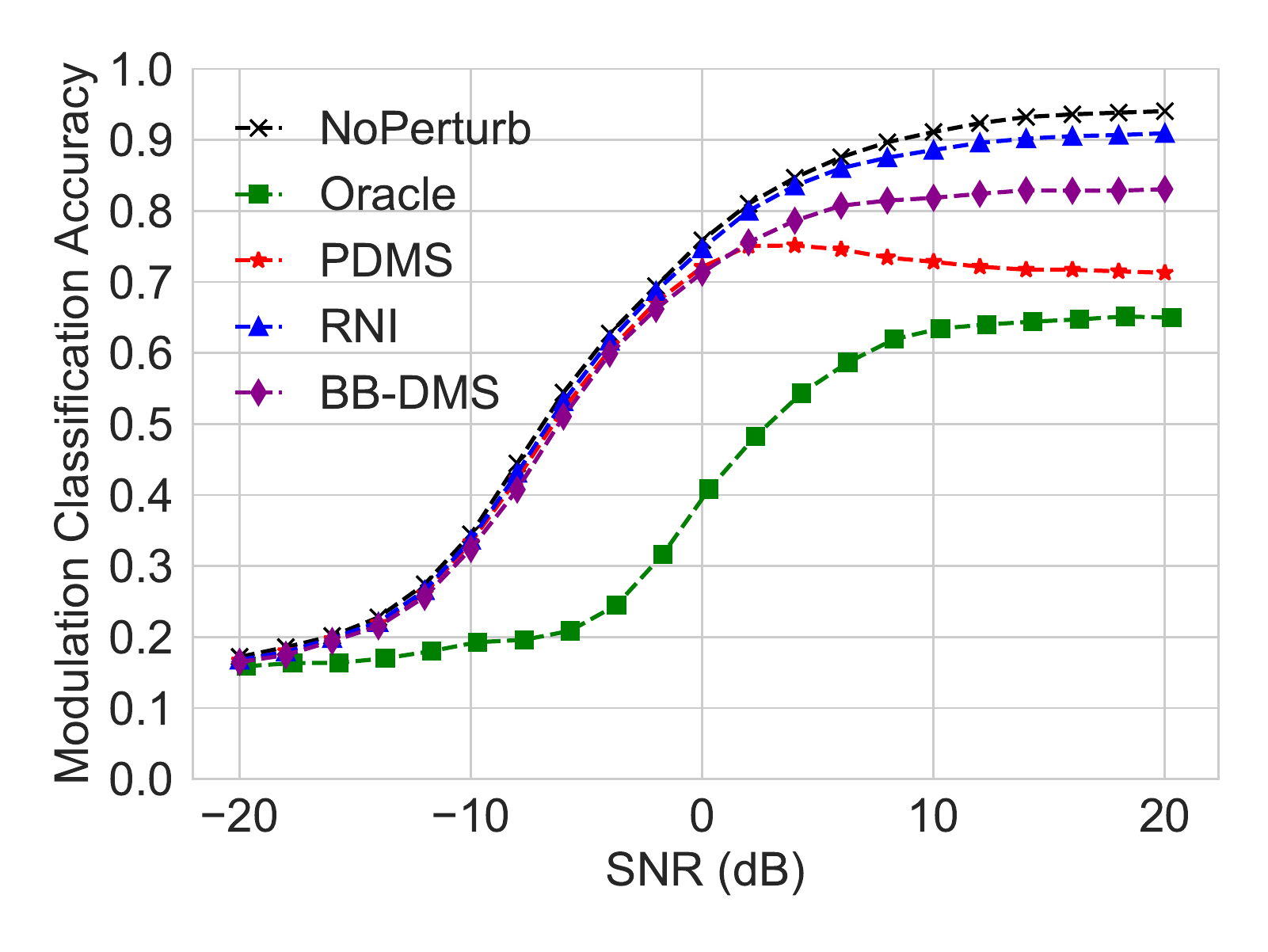}
     }

	\caption{Modulation-classification accuracy for an intruder trained with/without curriculum training for a complete dataset of channel SNR values ranging from -20 dB to 20 dB ($\epsilon$ = 3).}
     \label{fig: accuracy_mix_dnn_complete}
\end{figure}

\subsection{Improving intruder's performance by diversifying the training data}\label{sec: div_training}

In this section, we consider the scenario when the intruder has training data available at different SNR values ranging from -20 dB to 20 dB. We consider a larger training set consisting of training samples for diverse channel SNR values (21 different SNR levels uniformly spaced between -20 dB to 20 dB), leading to a total of 21 $\times$ 12966 samples. We consider two different training strategies: (i) randomly shuffle the training data of all channel SNR values to train the intruder's DNN; and (ii) curriculum learning \cite{bengio2009curriculum}, where the training data is arranged in descending order of their SNR values, and the training is started with samples of training data from SNR 20 dB, gradually adding samples with lower SNR values.

\begin{figure}[t]
\centering
 \subfloat[Without curriculum training. \label{subfig-2:qam64_mix_dnn_complete_noncur}]{%
       \includegraphics[width=\graphwidth]{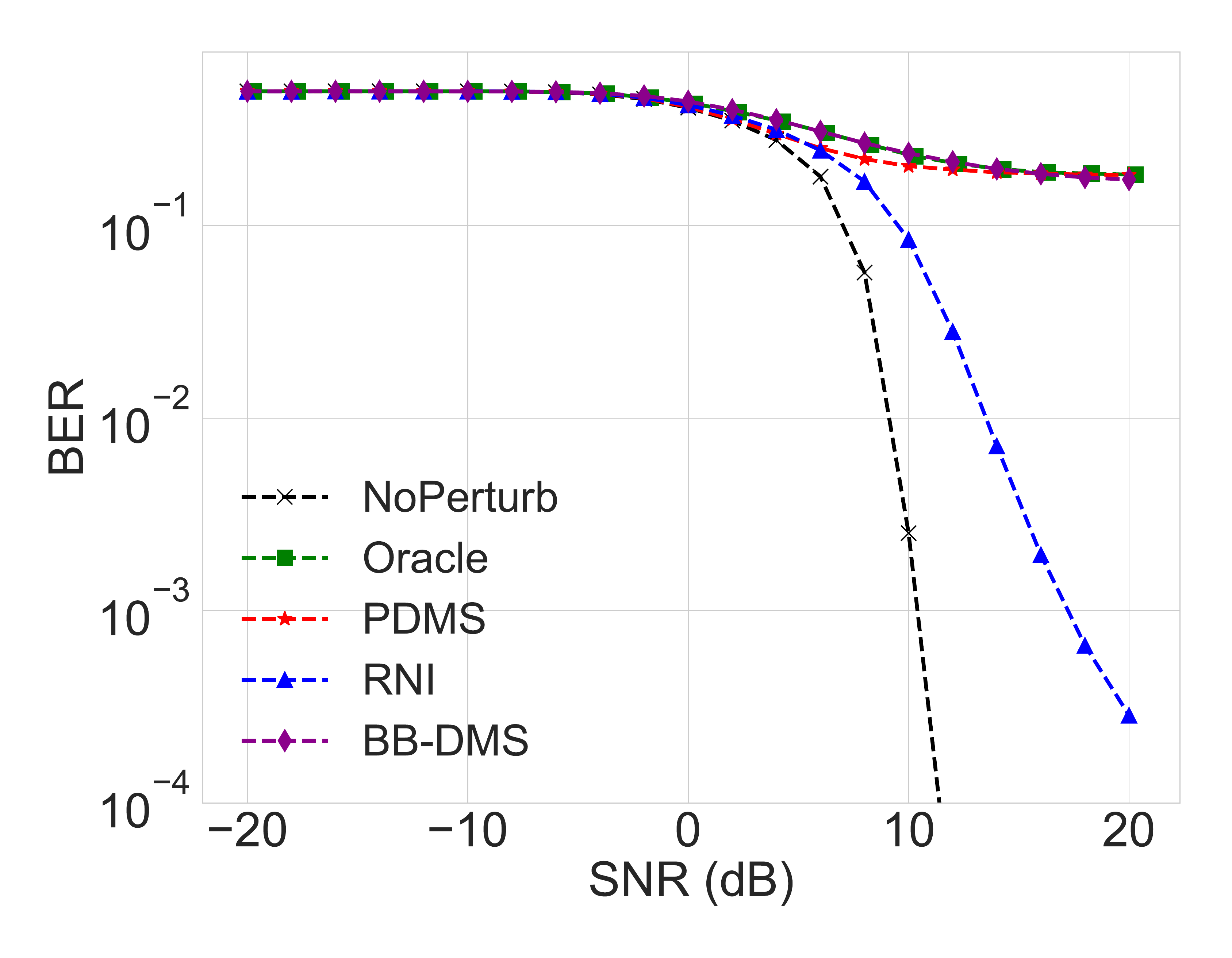}}
  \hfill
  \subfloat[With curriculum training. \label{subfig-2:qam64_mix_dnn_complete_cur}]{%
       \includegraphics[width=\graphwidth]{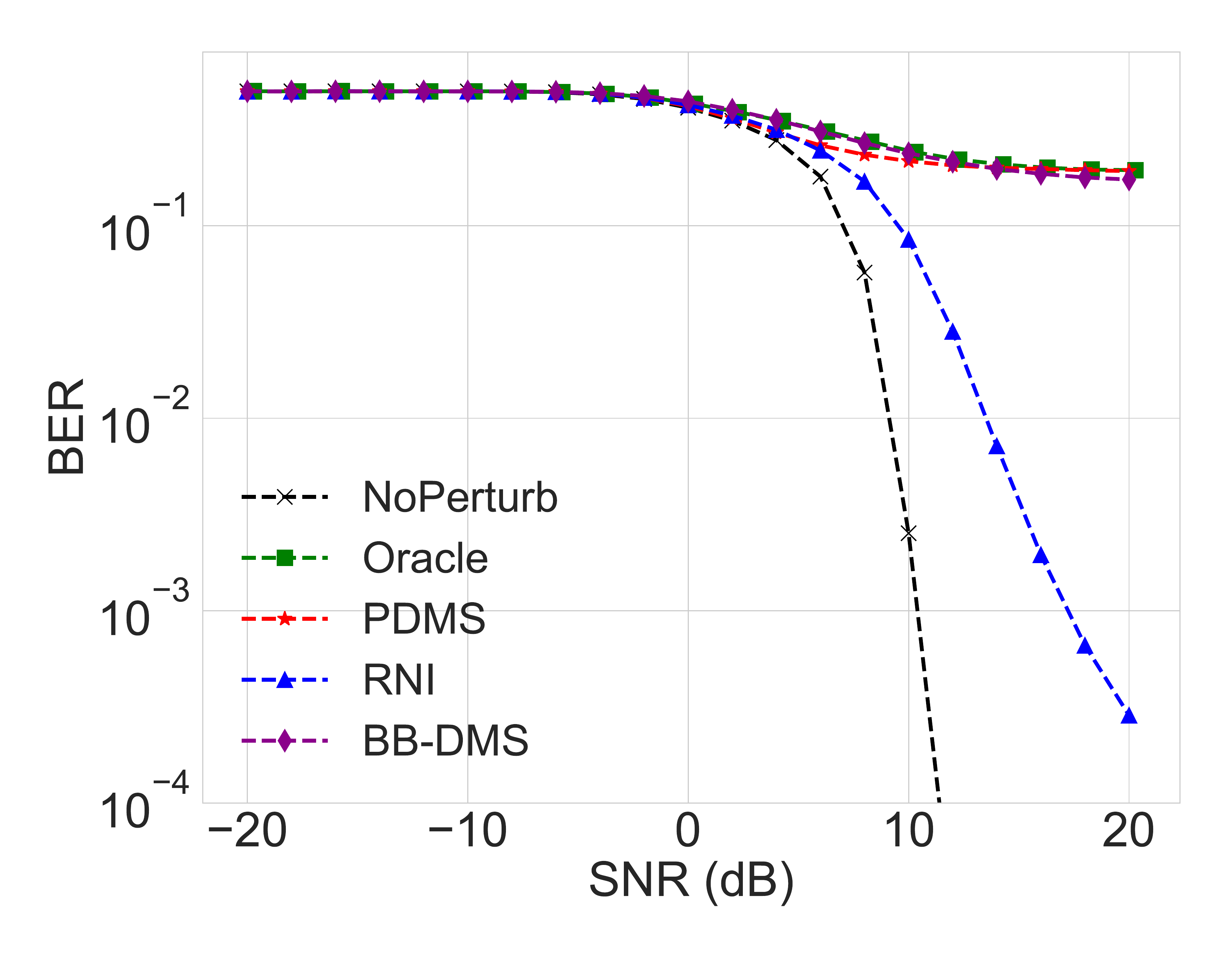}}
	\caption{BER for \QAM{64} for intruder trained with/without curriculum training with complete dataset of channel SNR values ranging from -20 dB to 20 dB ($\epsilon$ = 3).}
     \label{fig: ber_mix_dnn_complete}
\end{figure}

Fig.~\ref{subfig-1:accuracy_mix_dnn_complete_noncur} shows that an intruder network trained with data from all SNR values achieves a higher modulation-classification accuracy for \NoPerturb and against all defensive modulation strategies compared to the case when only samples from the same SNR values were used (cf. Fig.~\ref{fig: accu_cr_2_by_3}); this is most likely due to the approximately 20-fold increase in the number of training samples used. On the other hand, we can see in Fig.~\ref{subfig-1:accuracy_mix_dnn_complete_cur} that curriculum training achieves even higher robustness against all the defensive modulation schemes, and even the idealized defensive modulation scheme \Oracle can be detected with more than 60\% accuracy. This is because, in curriculum training, the neural network gradually learns, starting from easier concepts to more complex ones (more noisy channels in our case) and generalizes better to unseen data including those generated by defensive modulation schemes. In both cases, the improvement in detection accuracy is more for higher SNR values. Fig.~\ref{fig: ber_mix_dnn_complete} shows the BER for \QAM{64} when defensive perturbations are used against these intruder modulation classifiers trained over the whole range of SNR values without and with curriculum training, respectively. The achieved BERs are similar to those achieved when the intruder classifiers are trained for a particular SNR in Fig.~\ref{fig: ber_cr_2_by_3}. This shows that the comparison of the detection accuracy discussed above is fair (i.e., the improved detection accuracy is not because the applied defensive perturbations are smaller).

\begin{figure}[t]
\centering
   \subfloat[Modulation-classification accuracy  \label{subfig-1:spsa_accuracy_mixed_dnn_noncur}]{%
       \includegraphics[width=\graphwidth]{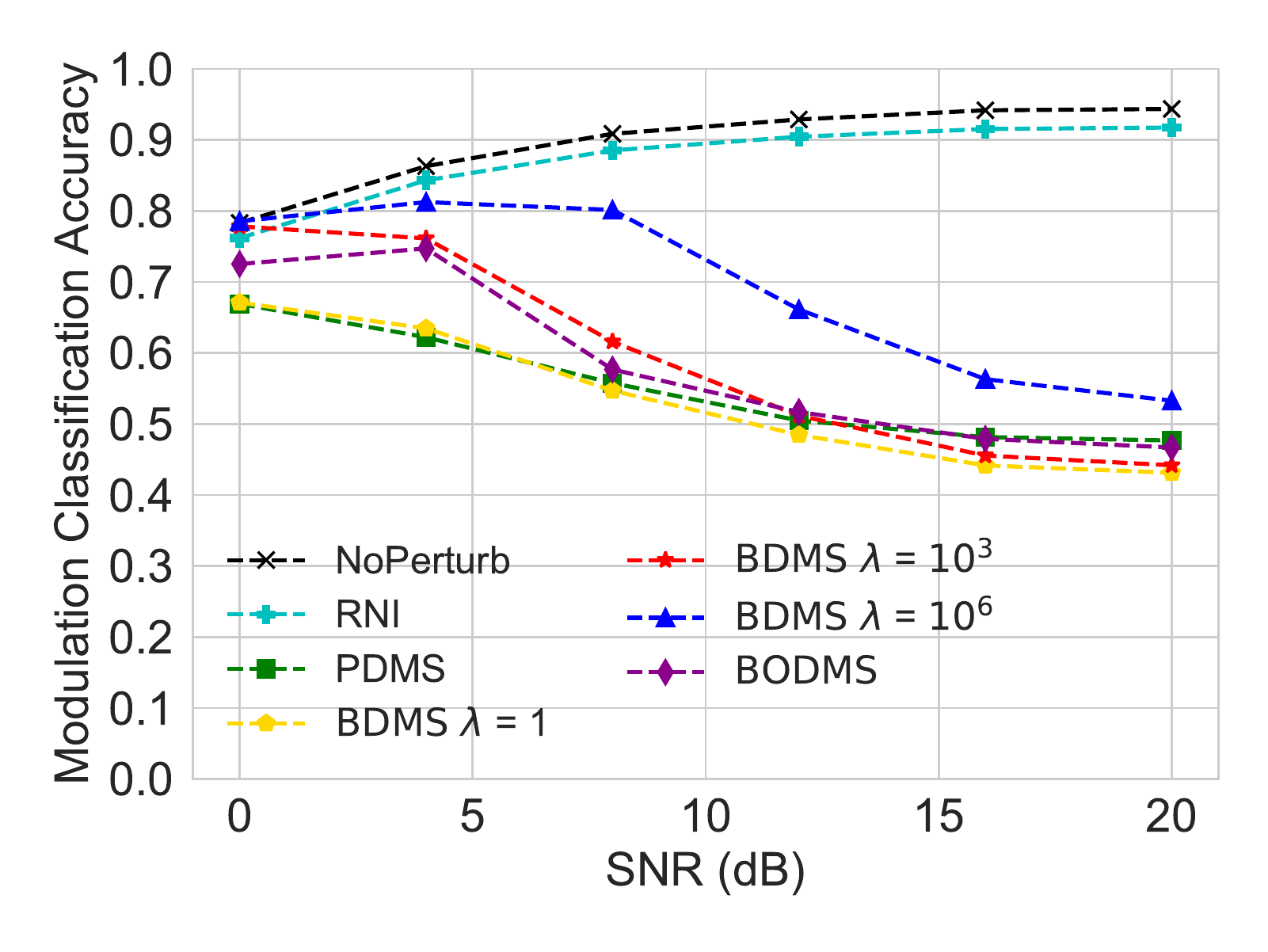}
     }
     \hfill
     \subfloat[BER for \QAM{64} \label{subfig-2:spsa_qam64_mixed_dnn_noncur}]{%
       \includegraphics[width=\graphwidth]{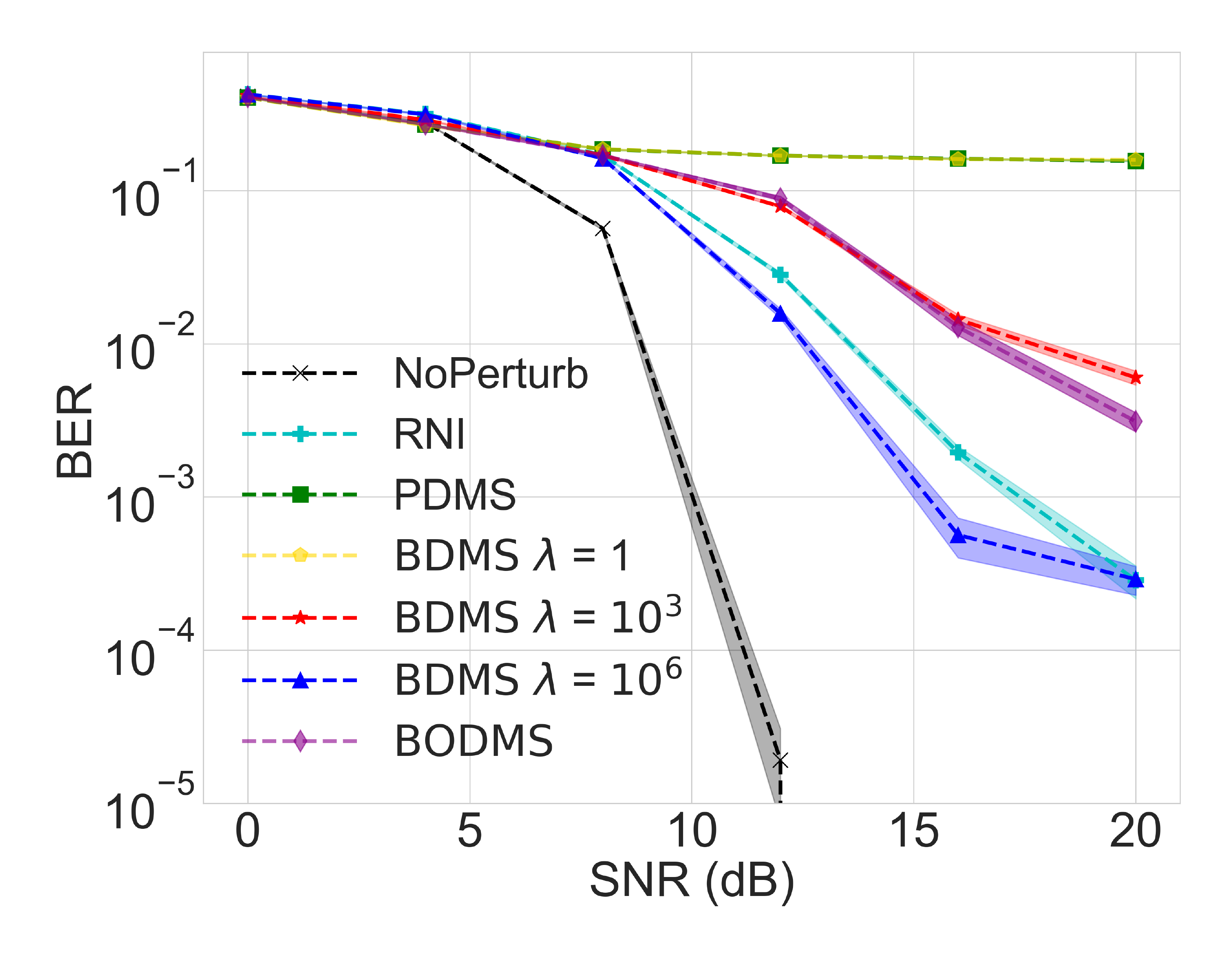}}
	\caption{DNN classifier trained with training data of channel SNRs -20 dB to 20 dB without any curriculum ($\epsilon$ = 3).}
     \label{fig: spsa_mix_dnn_noncur}
\end{figure}

Next, we consider the performance of BER-aware defensive modulation schemes when the intruder classifiers are trained with complete training data of all channel SNR values. The results without any curriculum learning are shown in Fig.~\ref{fig: spsa_mix_dnn_noncur} for the same DNN-based classifier.  It can be seen that the modulation-classification accuracy is quite high, around 95\%, when no defense mechanism is employed (\NoPerturb), and over 90\% when only noise is added (\RNI). We can also observe that, compared to results in Fig.~\ref{subfig-1:accuracy_mix_dnn_complete_noncur}, \BDMS is less successful against this model for large $\lambda$ ($10^6$); on the other hand, the BER is significantly improved, as demonstrated by comparing Fig.~\ref{subfig-2:qam64_mix_dnn_complete_noncur} and Fig.~\ref{subfig-2:spsa_qam64_mixed_dnn_noncur}. There is also a significant improvement in detection accuracy for essentially the same BER compared to the case when only training data for the same SNR value is used (cf. Fig.~\ref{fig: accuracy_spsa_dnn} and Fig.~\ref{fig: ber_spsa_dnn}).

On the other hand, using this larger set of training data yields no significant improvement in the performance of the tree-based classifier, and the results are very similar to those reported in Fig.~\ref{fig: spsa_tree} (hence, they are omitted).

\begin{figure}[t]
\centering
   \subfloat[Modulation-classification accuracy  \label{subfig-1:spsa_accuracy_mixed_dnn_cur}]{%
       \includegraphics[width=\graphwidth]{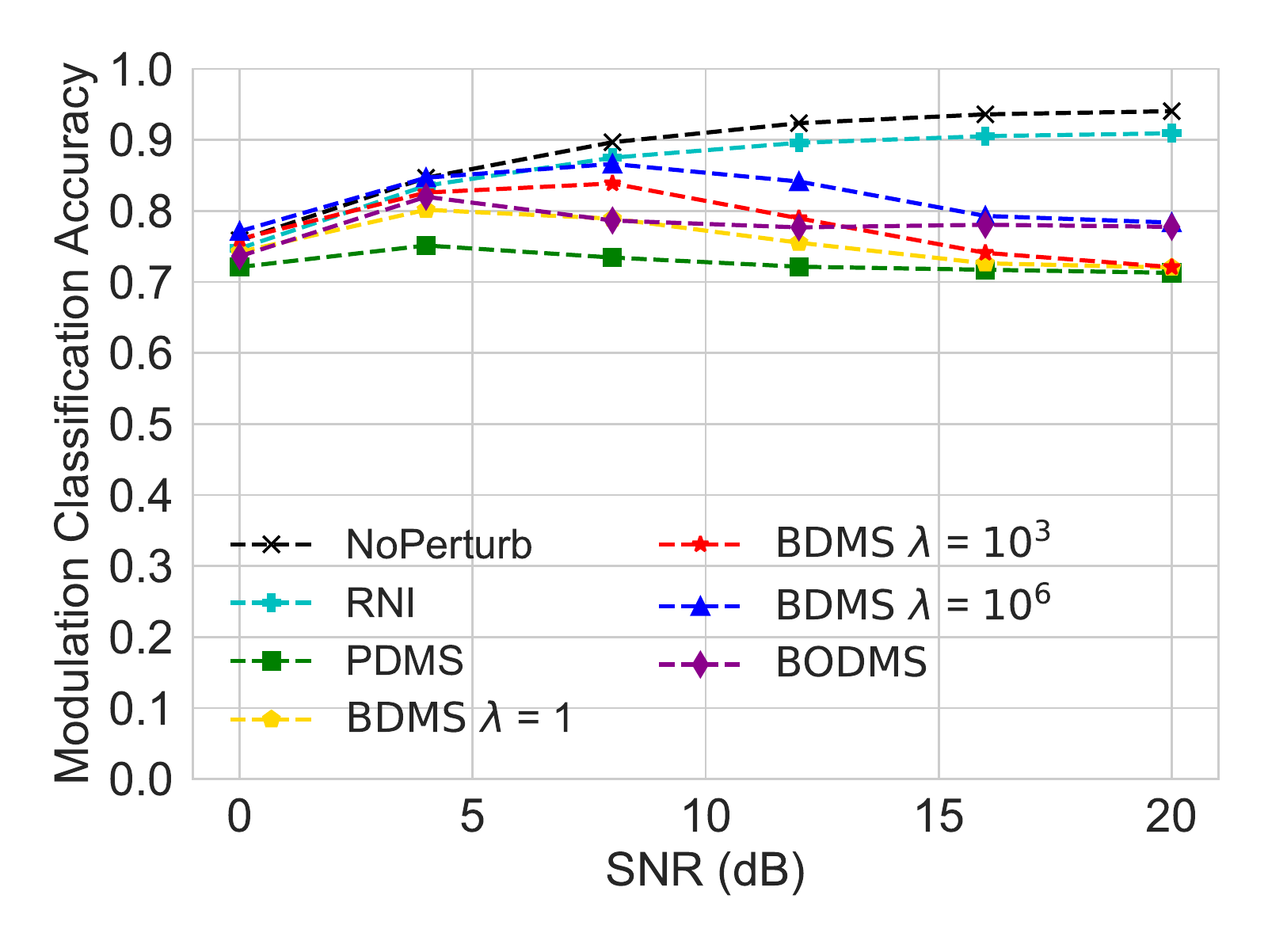}
     }
     \hfill
     \subfloat[BER for \QAM{64} \label{subfig-2:spsa_qam64_mixed_dnn_cur}]{%
       \includegraphics[width=\graphwidth]{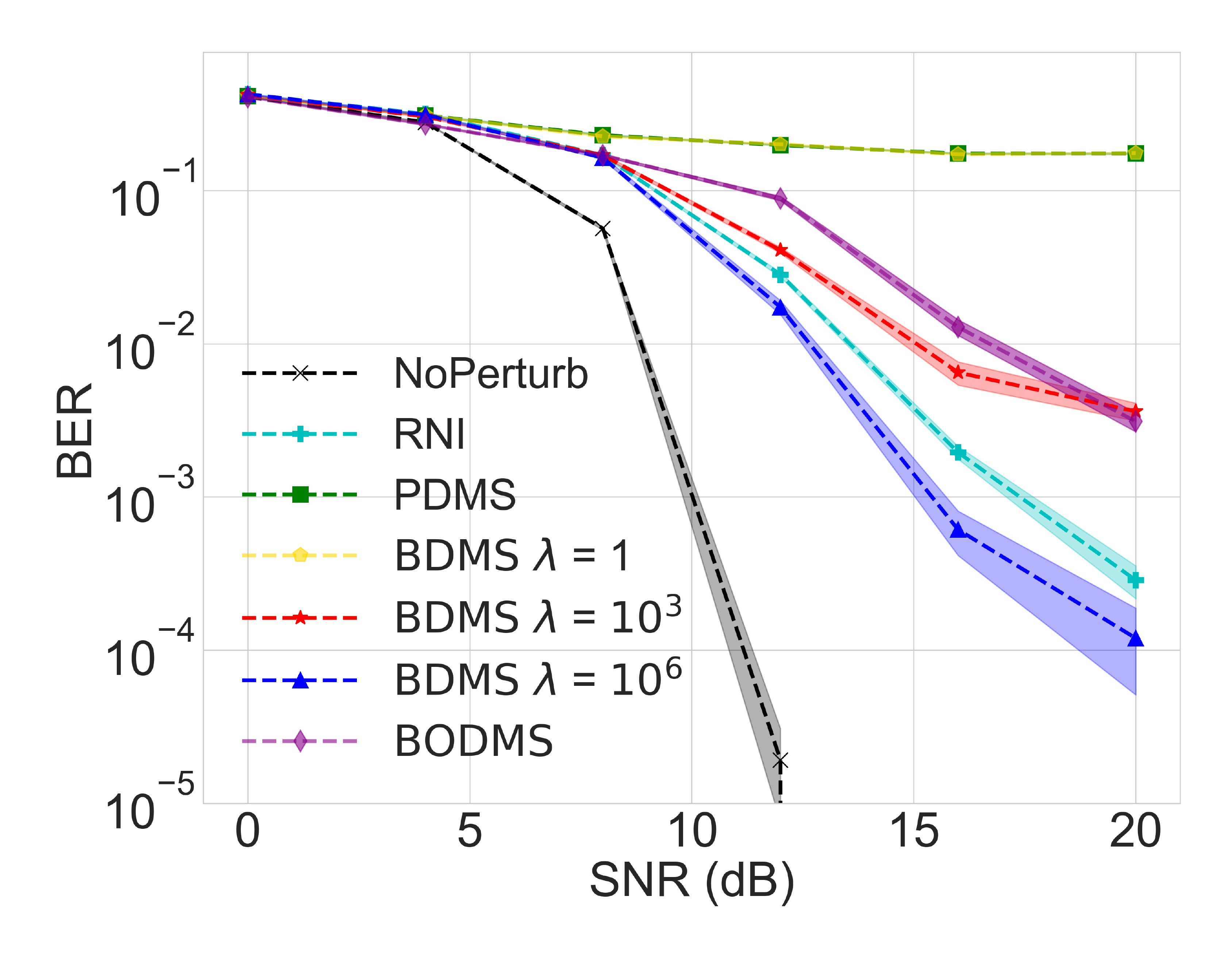}}
	\caption{DNN classifier trained with data of channel SNRs -20 dB to 20 dB with curriculum learning ($\epsilon$ = 3).}
     \label{fig: spsa_mix_dnn_cur}
\end{figure}

When the DNN-based classifier is trained using the complete dataset with curriculum learning, a significantly higher modulation-classification accuracy can be achieved against all defensive modulation schemes, as shown in  Fig.~\ref{fig: spsa_mix_dnn_cur}. Compared to the non-curriculum learning results in Fig.~\ref{fig: spsa_mix_dnn_noncur}, we can see that the improved detection accuracy also results in a smaller BER.
This suggests that, for a fair comparison between the two approaches, we can increase the attack strength in the case of curriculum learning until we achieve similar BER values as in Fig.~\ref{fig: spsa_mix_dnn_noncur}.

\begin{figure}[t]
\centering
   \subfloat[Modulation-classification accuracy  \label{subfig-1:accuracy_mix_dnn_curriculum_coderate_2_3_norm}]{%
       \includegraphics[width=\graphwidth]{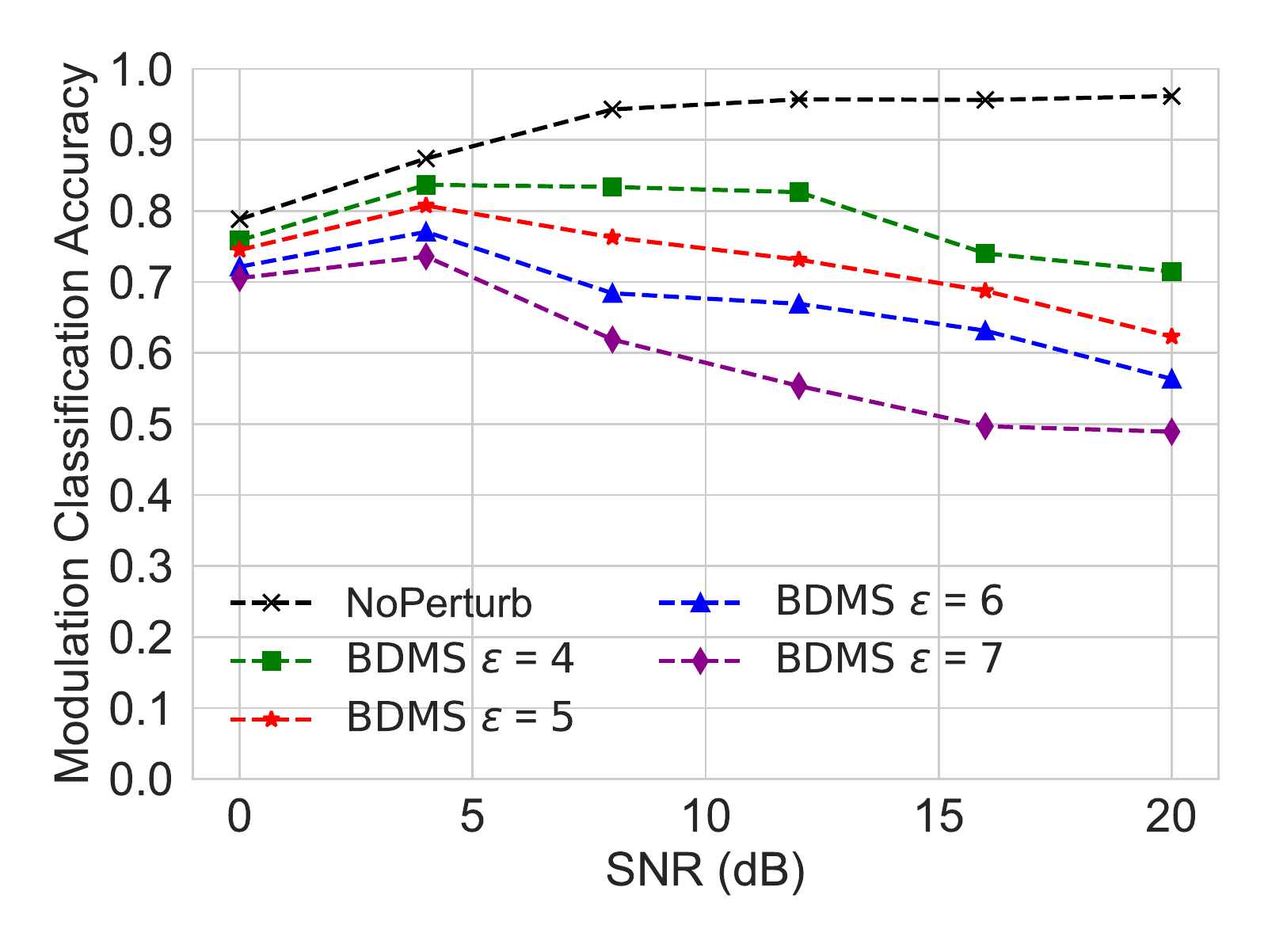}
     }
     \hfill
    \subfloat[BER for \QAM{64} \label{subfig-2:qam64_mix_dnn_curriculum_coderate_2_3_norm}]{%
       \includegraphics[width=\graphwidth]{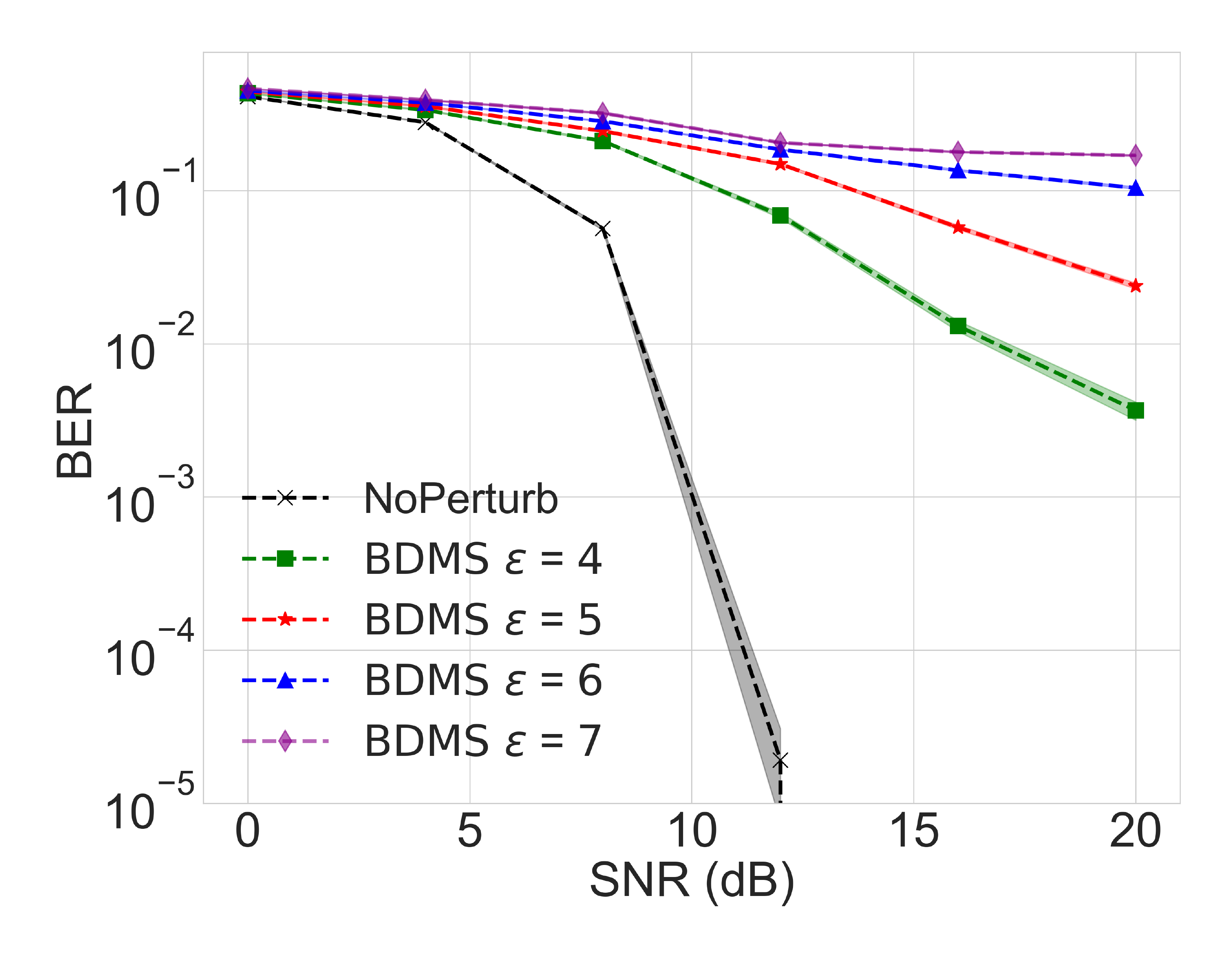}}


	\caption{Modulation-classification accuracy and BER (QAM{64}) for an intruder trained with a dataset of channel SNR ranging from -20 dB to 20 dB with curriculum learning (code rate = $2/3$, BDMS with $\lambda$ = $10^{20}$).}
     \label{fig: mix_dnn_curriculum_cr_2_3_norm}
\end{figure}

To this end, we increase the norm of perturbations for the \BDMS scheme against the DNN-based intruder network trained with curriculum learning. Note that to make the defense mechanisms work, we need to increase the value of $\lambda$, and we have found that (the surprisingly large) $\lambda = 10^{20}$ works well in our experiments. The results are shown in Fig.~\ref{fig: mix_dnn_curriculum_cr_2_3_norm}. It can be seen that defensive perturbations with larger norms decrease the modulation-detection accuracy of the intruder, but they also result in significantly higher BER despite the very large $\lambda$ value.

The results in this section showed that using more and diverse data and curriculum training can significantly improve the performance of the intruder and its robustness against various defense mechanisms. While designing better defense mechanisms against these intruders is an interesting and challenging future research direction, one method that can be employed directly at the transmitter is to reduce the code rate, which allows employing stronger attacks at the transmitter. This is explored in the next section.

\subsection{The effect of the code rate}

In our previous experiments we considered a fixed code rate of $2/3$. However, we have observed that the BER increases significantly for some of the modulation schemes due to input perturbations. One way to reduce the BER in the presence of the defensive perturbations to the transmitted signal is to introduce additional redundancy by decreasing the code rate.

\begin{figure}[t]
\centering
   \subfloat[Modulation-classification accuracy  \label{subfig-1:accuracy_spsa_dnn_1_2}]{%
       \includegraphics[width=\graphwidth]{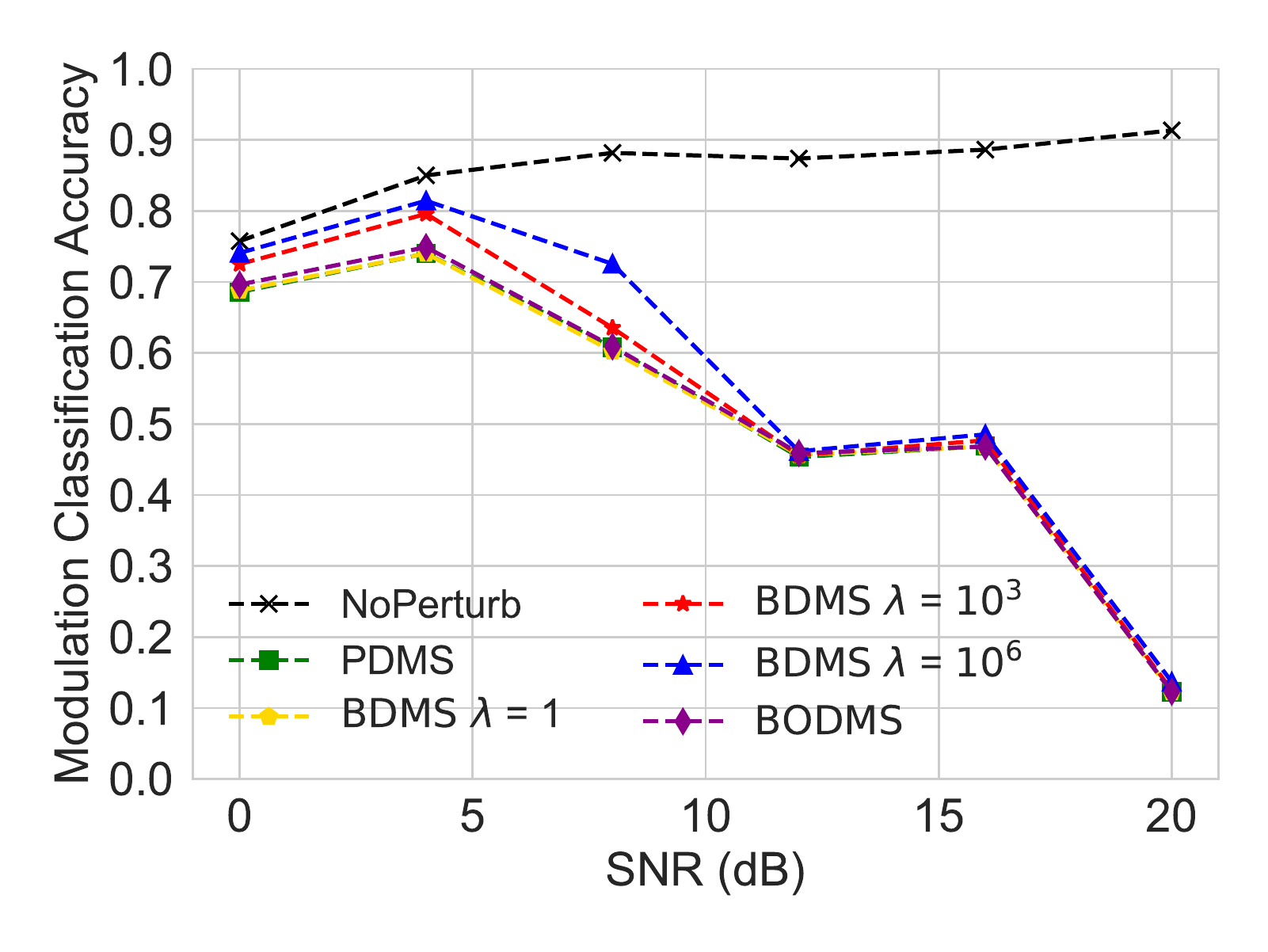}
     }
     \hfill
     \subfloat[BER for \QAM{64} \label{subfig-2:spsa_qam64_dnn_1_2}]{%
       \includegraphics[width=\graphwidth]{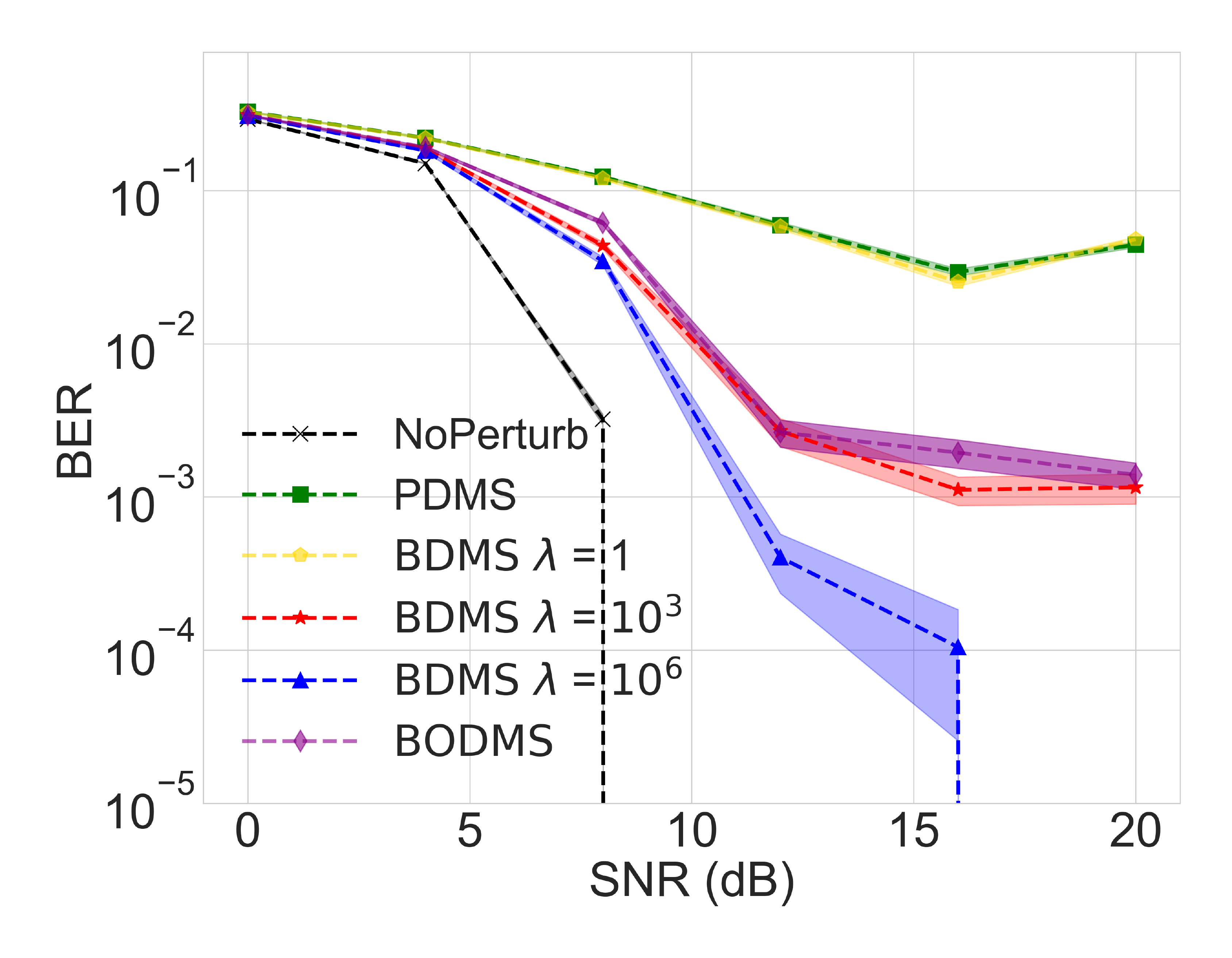}}
	\caption{Modulation-classification accuracy of DNN-based intruder and bit error rate (QAM{64}) for code rate $1/2$ for BER-aware modulation schemes ($\epsilon$ = 3).}
     \label{fig: spsa_dnn_1_2}
\end{figure}

To illustrate the effect of the code rate, we first evaluate the performance of our BER-aware defense schemes (with $\epsilon = 3$) for a channel code of rate $1/2$ against the usual DNN-based intruder trained for a specific SNR. The results, shown in Fig.~\ref{fig: spsa_dnn_1_2} demonstrate that both the BER and the detection accuracy can be substantially reduced compare to the case when the code rate is $2/3$ (see Fig.~\ref{fig: accuracy_spsa_dnn} and Fig.~\ref{fig: ber_spsa_dnn} for comparison). For example, even for \QAM{64}, \BDMS (with $\lambda= 10^6$) achieves zero BER for high SNR values (at least 16 dB).

\begin{figure}[t]
\centering
   \subfloat[Modulation-classification accuracy  \label{subfig-1:accuracy_mix_dnn_curriculum_coderate_1_2_norm}]{%
       \includegraphics[width=\graphwidth]{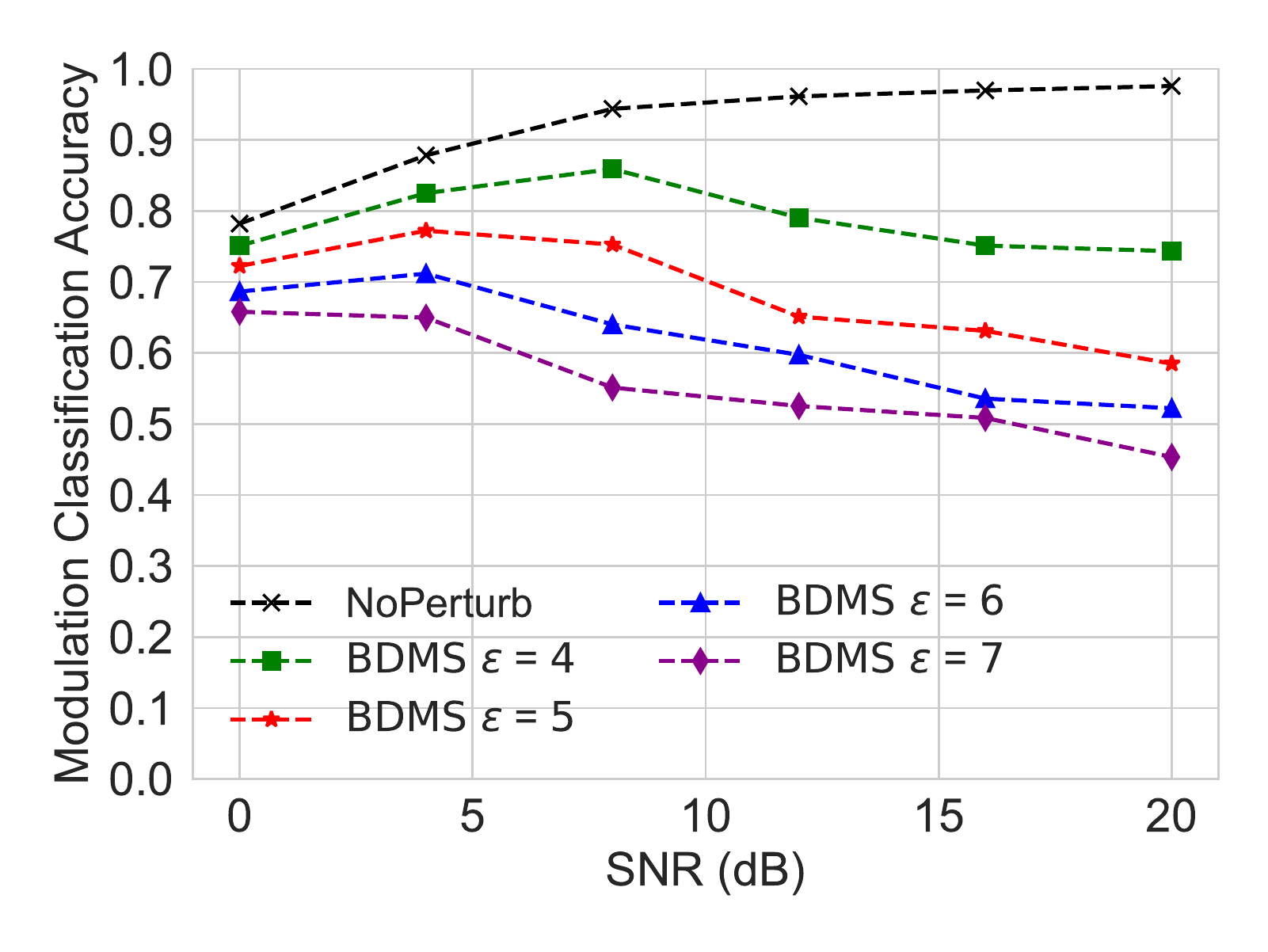}
     }
     \hfill
    \subfloat[BER for \QAM{64} \label{subfig-2:qam64_mix_dnn_curriculum_coderate_1_2_norm}]{%
       \includegraphics[width=\graphwidth]{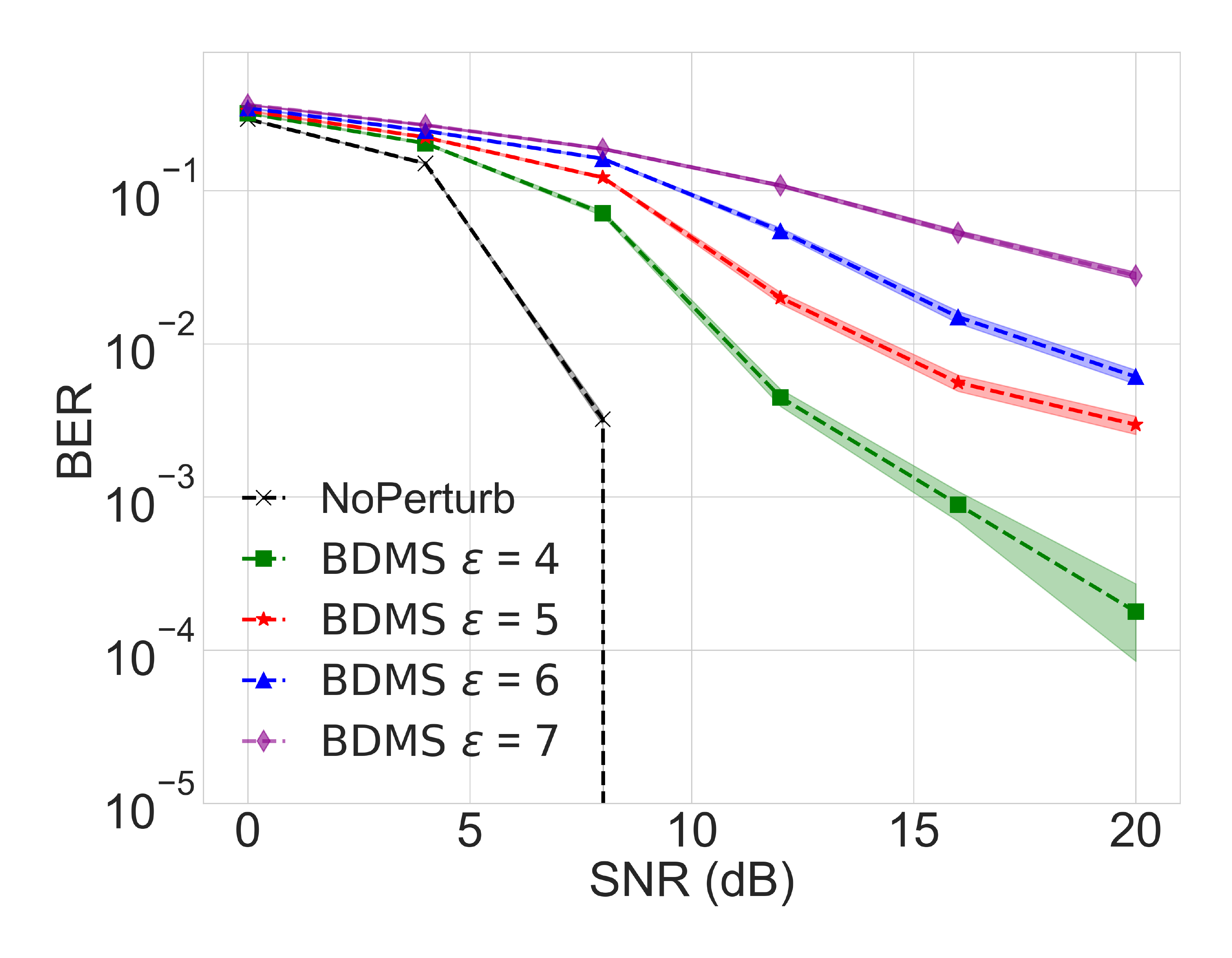}}

	\caption{Modulation-classification accuracy and BER (QAM{64}) for an intruder trained with a dataset of channel SNR ranging from -20 dB to 20 dB with curriculum learning (code rate = $1/2$, BDMS with $\lambda$ = $10^{20}$).}
     \label{fig: mix_dnn_curriculum_cr_1_2_norm}
\end{figure}

The very small BERs (obtained in the previous experiment) allow the application of more aggressive
defensive perturbations when the intruder employs a stronger classifier. Accordingly, we evaluate the \BDMS defensive scheme (with a large $\lambda =10^{20}$) for different perturbation norms against a DNN-based intruder trained with curriculum learning over a range of SNR values (the setup is the same as for Fig.~\ref{fig: spsa_mix_dnn_cur} except for the code rate). The results, shown in Fig.~\ref{fig: mix_dnn_curriculum_cr_1_2_norm}, demonstrate that, compared to Fig.~\ref{fig: mix_dnn_curriculum_cr_2_3_norm}, using a lower code rate of $1/2$, the modulation-classification accuracy of the intruder trained with curriculum learning can be reduced without incurring a large BER at the legitimate receiver.

\section{Conclusions and Future Work}
\label{sec:conclusions}

We proposed a novel approach to secure wireless communication by preventing an intruder
from detecting the modulation scheme employed, which is typically the first step of a more advanced attack. In the proposed scheme, the I/Q symbols of the modulated waveform at the transmitter are perturbed using an adversarial perturbation derived against the modulation classifier of the intruder. The perturbation is designed using PGD, whose goal is to identify a perturbation with a limited norm that is sufficient to fool the intruder's classifier. More advanced methods are also proposed, whose goal is also to keep small the BER caused by the perturbation at the legitimate receiver. Experimental results verify the viability of our approach by showing that our methods are able to substantially reduce the modulation-classification accuracy of the intruder with minimal sacrifice in the communication performance. We have also shown that the intruder can improve its detection accuracy significantly by training with a dataset of samples taken from a range of SNR values, especially when curriculum learning is also employed. This provides robustness against channel noise as well as potential defense mechanisms against the intruder, and has led to improvements upon state-of-the-art modulation detectors in our experiments. Finally we have shown that a better trade-off between the intruder's detection accuracy and the BER at the legitimate receiver can be achieved by sacrificing the communication rate.

Utilizing the fast advances in the field of adversarial machine learning, our defense methods can certainly be improved in the future by applying more advanced as well as more universal (e.g., black box) adversarial attack methods. Another interesting avenue for future research is to develop sophisticated defensive perturbations that can exploit different channel characteristics both at the intruder and legitimate receiver. On the other end of the problem, one can develop better training strategies for the intruder that can achieve more robust performance against these defense mechanisms, for example, by applying adversarial training methods \cite{madry2017towards}.

\bibliographystyle{IEEEtran}
\bibliography{extended_arXiv_version}

\begin{thebibliography}{10}
\providecommand{\url}[1]{#1}
\csname url@samestyle\endcsname
\providecommand{\newblock}{\relax}
\providecommand{\bibinfo}[2]{#2}
\providecommand{\BIBentrySTDinterwordspacing}{\spaceskip=0pt\relax}
\providecommand{\BIBentryALTinterwordstretchfactor}{4}
\providecommand{\BIBentryALTinterwordspacing}{\spaceskip=\fontdimen2\font plus
\BIBentryALTinterwordstretchfactor\fontdimen3\font minus
  \fontdimen4\font\relax}
\providecommand{\BIBforeignlanguage}[2]{{%
\expandafter\ifx\csname l@#1\endcsname\relax
\typeout{** WARNING: IEEEtran.bst: No hyphenation pattern has been}%
\typeout{** loaded for the language `#1'. Using the pattern for}%
\typeout{** the default language instead.}%
\else
\language=\csname l@#1\endcsname
\fi
#2}}
\providecommand{\BIBdecl}{\relax}
\BIBdecl

\bibitem{hameed2019communicationGS}
M.~Z. Hameed, A.~Gy{\"o}rgy, and D.~G{\"u}nd{\"u}z, ``Communication without
  interception: Defense against modulation detection,'' in \emph{2019 {IEEE}
  Global Conference on Signal and Information Processing (GlobalSIP 2019)},
  Ottawa, ON, Canada, November 2019.

\bibitem{Prescott1993PerformanceMF}
G.~E. Prescott, ``Performance metrics for low probability of
  intercept-communication system,'' in \emph{Air Force Off. of Sci. Res., Tech.
  Rep.}, 1993.

\bibitem{Wyner:BSTJ:75}
A.~D. Wyner, ``The wire-tap channel,'' \emph{The Bell Sys. Tech. Journal},
  vol.~54, no.~8, pp. 1355--1387, Oct 1975.

\bibitem{Gunduz:security_feedback}
D.~Gunduz, D.~R. Brown, and H.~V. Poor, ``Secret communication with feedback,''
  in \emph{2008 International Symposium on Information Theory and Its
  Applications}.\hskip 1em plus 0.5em minus 0.4em\relax IEEE, 2008, pp. 1--6.

\bibitem{Bash:arxiv:12}
B.~A. Bash, D.~Goeckel, and D.~Towsley, ``Square root law for communication
  with low probability of detection on awgn channels,'' in \emph{2012 IEEE
  International Symposium on Information Theory Proceedings}.\hskip 1em plus
  0.5em minus 0.4em\relax IEEE, 2012, pp. 448--452.

\bibitem{dobre2007survey}
O.~A. Dobre, A.~Abdi, Y.~Bar-Ness, and W.~Su, ``Survey of automatic modulation
  classification techniques: classical approaches and new trends,'' \emph{IET
  communications}, vol.~1, no.~2, pp. 137--156, 2007.

\bibitem{mendis2016deep}
G.~J. Mendis, J.~Wei, and A.~Madanayake, ``Deep learning-based automated
  modulation classification for cognitive radio,'' in \emph{2016 IEEE
  International Conference on Communication Systems (ICCS)}.\hskip 1em plus
  0.5em minus 0.4em\relax IEEE, 2016, pp. 1--6.

\bibitem{o2017introduction}
T.~O’Shea and J.~Hoydis, ``An introduction to deep learning for the physical
  layer,'' \emph{IEEE Transactions on Cognitive Communications and Networking},
  vol.~3, no.~4, pp. 563--575, 2017.

\bibitem{west2017deep}
N.~E. West and T.~O'Shea, ``Deep architectures for modulation recognition,'' in
  \emph{2017 IEEE International Symposium on Dynamic Spectrum Access Networks
  (DySPAN)}.\hskip 1em plus 0.5em minus 0.4em\relax IEEE, 2017, pp. 1--6.

\bibitem{kim2016deep}
B.~Kim, J.~Kim, H.~Chae, D.~Yoon, and J.~W. Choi, ``Deep neural network-based
  automatic modulation classification technique,'' in \emph{2016 International
  Conference on Information and Communication Technology Convergence
  (ICTC)}.\hskip 1em plus 0.5em minus 0.4em\relax IEEE, 2016, pp. 579--582.

\bibitem{liu2017deep}
X.~Liu, D.~Yang, and A.~El~Gamal, ``Deep neural network architectures for
  modulation classification,'' in \emph{51st Asilomar Conference on Signals,
  Systems and Computers}, 2017.

\bibitem{szegedy2013intriguing}
J.~Bruna, C.~Szegedy, I.~Sutskever, I.~Goodfellow, W.~Zaremba, R.~Fergus, and
  D.~Erhan, ``Intriguing properties of neural networks,'' in
  \emph{International Conference on Learning Representations}, 2014.

\bibitem{goodfellow2014explaining}
I.~J. Goodfellow, J.~Shlens, and C.~Szegedy, ``Explaining and harnessing
  adversarial examples,'' in \emph{International Conference on Learning
  Representations}, 2015.

\bibitem{sadeghi2018adversarial}
M.~Sadeghi and E.~G. Larsson, ``Adversarial attacks on deep-learning based
  radio signal classification,'' \emph{IEEE Wireless Comm. Letters}, 2018.

\bibitem{kokalj2019mitigation}
S.~Kokalj-Filipovic, R.~Miller, N.~Chang, and C.~L. Lau, ``Mitigation of
  adversarial examples in rf deep classifiers utilizing autoencoder
  pre-training,'' in \emph{2019 International Conference on Military
  Communications and Information Systems (ICMCIS)}.\hskip 1em plus 0.5em minus
  0.4em\relax IEEE, 2019, pp. 1--6.

\bibitem{kokalj2019targeted}
S.~Kokalj-Filipovic, R.~Miller, and J.~Morman, ``Targeted adversarial examples
  against {RF} deep classifiers,'' in \emph{Proceedings of the ACM Workshop on
  Wireless Security and Machine Learning}.\hskip 1em plus 0.5em minus
  0.4em\relax ACM, 2019, pp. 6--11.

\bibitem{carlini2017towards}
N.~Carlini and D.~Wagner, ``Towards evaluating the robustness of neural
  networks,'' in \emph{2017 IEEE Symposium on Security and Privacy (SP)}.\hskip
  1em plus 0.5em minus 0.4em\relax IEEE, 2017, pp. 39--57.

\bibitem{hameed2019communication}
M.~Z. Hameed, A.~Gy\"orgy, and D.~G\"und\"uz, ``Communication without
  interception: Defense against deep-learning-based modulation detection,''
  \emph{arXiv preprint arXiv:1902.10674}, 2019.

\bibitem{flowers2019communications}
B.~Flowers, R.~M. Buehrer, and W.~C. Headley, ``Communications aware
  adversarial residual networks for over the air evasion attacks,'' in
  \emph{MILCOM 2019-2019 IEEE Military Communications Conference
  (MILCOM)}.\hskip 1em plus 0.5em minus 0.4em\relax IEEE, 2019, pp. 133--140.

\bibitem{flowers2019evaluating}
B.~{Flowers}, R.~M. {Buehrer}, and W.~C. {Headley}, ``Evaluating adversarial
  evasion attacks in the context of wireless communications,'' \emph{IEEE
  Transactions on Information Forensics and Security}, vol.~15, pp. 1102--1113,
  2020.

\bibitem{kim2020overtheair}
B.~Kim, Y.~E. Sagduyu, K.~Davaslioglu, T.~Erpek, and S.~Ulukus, ``Over-the-air
  adversarial attacks on deep learning based modulation classifier over
  wireless channels,'' \emph{arXiv preprint arXiv:2002.02400}, 2020.

\bibitem{barreno2010taxonomy}
M.~Barreno, B.~Nelson, A.~D. Joseph, and J.~D. Tygar, ``The security of machine
  learning,'' \emph{Machine Learning}, vol.~81, no.~2, pp. 121--148, 2010.

\bibitem{sagduyu2019iot}
Y.~E. Sagduyu, Y.~Shi, and T.~Erpek, ``{IoT} network security from the
  perspective of adversarial deep learning,'' in \emph{2019 16th Annual IEEE
  International Conference on Sensing, Communication, and Networking
  (SECON)}.\hskip 1em plus 0.5em minus 0.4em\relax IEEE, 2019, pp. 1--9.

\bibitem{shi2018spectrum}
Y.~Shi, T.~Erpek, Y.~E. Sagduyu, and J.~H. Li, ``Spectrum data poisoning with
  adversarial deep learning,'' in \emph{MILCOM 2018-2018 IEEE Military
  Communications Conference (MILCOM)}.\hskip 1em plus 0.5em minus 0.4em\relax
  IEEE, 2018, pp. 407--412.

\bibitem{erpek2018deep}
T.~Erpek, Y.~E. Sagduyu, and Y.~Shi, ``Deep learning for launching and
  mitigating wireless jamming attacks,'' \emph{IEEE Transactions on Cognitive
  Communications and Networking}, vol.~5, no.~1, pp. 2--14, 2018.

\bibitem{kurakin2016adversarial}
A.~Kurakin, I.~Goodfellow, and S.~Bengio, ``Adversarial machine learning at
  scale,'' in \emph{International Conference on Learning Representations},
  2017.

\bibitem{chen2017ead}
P.-Y. Chen, Y.~Sharma, H.~Zhang, J.~Yi, and C.-J. Hsieh, ``Ead: elastic-net
  attacks to deep neural networks via adversarial examples,'' in
  \emph{Thirty-Second AAAI Conference on Artificial Intelligence}, 2018.

\bibitem{madry2017towards}
A.~Madry, A.~Makelov, L.~Schmidt, D.~Tsipras, and A.~Vladu, ``Towards deep
  learning models resistant to adversarial attacks,'' in \emph{International
  Conference on Learning Representations}, 2018.

\bibitem{papernot2017practical}
N.~Papernot, P.~McDaniel, I.~Goodfellow, S.~Jha, Z.~B. Celik, and A.~Swami,
  ``Practical black-box attacks against machine learning,'' in
  \emph{Proceedings of the 2017 ACM on Asia conference on computer and
  communications security}, 2017, pp. 506--519.

\bibitem{uesato2018adversarial}
J.~Uesato, B.~O’Donoghue, P.~Kohli, and A.~Oord, ``Adversarial risk and the
  dangers of evaluating against weak attacks,'' in \emph{International
  Conference on Machine Learning}, 2018, pp. 5025--5034.

\bibitem{spall1992multivariate}
J.~C. Spall, ``Multivariate stochastic approximation using a simultaneous
  perturbation gradient approximation,'' \emph{IEEE Trans. on Automatic Ctrl.},
  vol.~37, no.~3, pp. 332--341, 1992.

\bibitem{papernot2018cleverhans-orig}
N.~Papernot, F.~Faghri, N.~Carlini, I.~Goodfellow, R.~Feinman, A.~Kurakin,
  C.~Xie, Y.~Sharma, T.~Brown, A.~Roy, A.~Matyasko, V.~Behzadan,
  K.~Hambardzumyan, Z.~Zhang, Y.-L. Juang, Z.~Li, R.~Sheatsley, A.~Garg,
  J.~Uesato, W.~Gierke, Y.~Dong, D.~Berthelot, P.~Hendricks, J.~Rauber, and
  R.~Long, ``Technical report on the cleverhans v2.1.0 adversarial examples
  library,'' \emph{arXiv preprint arXiv:1610.00768}, 2018.

\bibitem{papernot2016transferability}
N.~Papernot, P.~McDaniel, and I.~Goodfellow, ``Transferability in machine
  learning: from phenomena to black-box attacks using adversarial samples,''
  \emph{arXiv preprint arXiv:1605.07277}, 2016.

\bibitem{Rosti98statisticalmethods}
A.-V. Rosti, ``Statistical methods in modulation classification,'' 1998.

\bibitem{abdelmutalab2016automatic}
A.~{Abdelmutalab \emph{et al.}}, ``Automatic modulation classification based on
  high order cumulants and hierarchical polynomial classifiers,''
  \emph{Physical Comm.}, vol.~21, pp. 10--18, 2016.

\bibitem{bengio2009curriculum}
Y.~Bengio, J.~Louradour, R.~Collobert, and J.~Weston, ``Curriculum learning,''
  in \emph{Proceedings of the 26th International Conference on Machine Learning
  (ICML)}.\hskip 1em plus 0.5em minus 0.4em\relax ACM, 2009, pp. 41--48.

\end{thebibliography}

\end{document}